\documentclass{article}

\usepackage{arxiv}

\usepackage[utf8]{inputenc} 
\usepackage[T1]{fontenc}    
\usepackage{hyperref}       
\usepackage{url}            
\usepackage{booktabs}       
\usepackage{amsfonts}       
\usepackage{nicefrac}       
\usepackage{microtype}      
\usepackage{lipsum}
\usepackage{graphicx}
\graphicspath{ {./images/} }
\pdfoutput=1

\usepackage{amsmath,amssymb} 
\usepackage{subfigure}
\usepackage{enumitem}
\usepackage{tikz}
\usepackage{hyperref}  
\usepackage{wrapfig}   
\usepackage{color} 
\usepackage{floatrow}
\newfloatcommand{capbtabbox}{table}[][\FBwidth]
\usepackage{booktabs} 
\usepackage{colortbl}
\usepackage{multirow}
\usepackage[switch]{lineno}

\newcommand\textblue[1]{\textcolor{blue}{#1}}

\hypersetup{
    colorlinks=true,
    linkcolor=red,
    filecolor=magenta,      
    urlcolor=cyan,
    pdftitle={Overleaf Example},
    pdfpagemode=FullScreen,
}

\title{Parallel Residual Bi-Fusion Feature Pyramid Network for Accurate Single-Shot Object Detection}

\author{
 Ping-Yang Chen \\
  Department of Computer Science\\
  National Yang Ming Chiao Tung University, Taiwan\\
  \texttt{pingyang.cs08@nycu.edu.tw} \\
   \And
 Ming-Ching Chang \\
  Department of Computer Science\\
  University at Albany, SUNY\\
  \texttt{mchang2@albany.edu} \\
  \And
 Jun-Wei Hsieh* \\
  College of Artificial Intelligence and Green Energy\\
  National Yang Ming Chiao Tung University, Taiwan\\
  \texttt{jwhsieh@nycu.edu.tw} \\
   \And
 Yong-Sheng Chen \\
  Department of Computer Science\\
  National Yang Ming Chiao Tung University, Taiwan\\
  \texttt{yschen@nycu.edu.tw} \\
}

\begin{document}
\maketitle
\begin{abstract}
This paper proposes the Parallel Residual Bi-Fusion Feature Pyramid Network (PRB-FPN) for fast and accurate single-shot object detection.
Feature Pyramid (FP) is widely used in recent visual detection, however the top-down pathway of FP cannot preserve accurate localization due to pooling shifting. The advantage of FP is weakened as deeper backbones with more layers are used.  In addition, it cannot  keep  up accurate detection of both small and large objects at the same time. 
To address these issues, we propose a new parallel FP structure with bi-directional (top-down and bottom-up) fusion and associated improvements to retain high-quality features for accurate localization.  We provide the following design improvements: (1) A parallel bifusion FP structure with a bottom-up fusion module (BFM) to detect both small and large objects at once with high accuracy. (2) A concatenation and re-organization (CORE) module provides a bottom-up pathway for feature fusion, which leads to the bi-directional fusion FP that can recover lost information from lower-layer feature maps. (3) The CORE feature is further purified to retain richer contextual information. Such CORE purification in both top-down and bottom-up pathways can be finished in only a few iterations. (4) The adding of a residual design to CORE leads to a new Re-CORE module that enables easy training and integration with a wide range of deeper or lighter backbones. The proposed network achieves state-of-the-art performance on the UAVDT17 and MS COCO datasets. Code is available at \textblue{\href{https://github.com/pingyang1117/PRBNet\_PyTorch}{https://github.com/pingyang1117/PRBNet\_PyTorch}}
\end{abstract}


\section{Introduction}
\label{sec:intro}

Visual object detection has improved significantly in the state-of-the-art (SoTA) models. Recent deep models including FPN~\cite{FPN}, YOLOv3~\cite{YOLOv3}, and SSD~\cite{SSD} typically consist of three components: (1) a deep {\em feature extraction backbone} {\em e.g.} DarkNet-53~\cite{YOLOv2} or ResNet-101~\cite{ResNet}, (2) a {\em feature pyramid} (FP), and (3) an {\em object classifier}. 
To ensure high detection accuracy, most SoTA object detectors adopt deep CNN structures that can achieve impressive performance in detecting large and medium sized objects. However the performance for detecting smaller objects are still inferior~\cite{LRF_Wang_2019_ICCV}. This is mainly because the feature map resolution is reduced after simple pooling in the FP. Tiny objects ($<32\times32$ pixels) can turn into about a single-pixel feature vector in the last layer of FP, causing insufficient spatial resolution for accurate discrimination.
On the other hand, using a shallow backbone increases the computational efficiency. This comes with the drawback of reduced detection performance, as the capability to retain contextual and semantic features also decreases directly. 

In general, detecting small objects is more difficult than detecting large objects. Both high-level and low-level features are required to discriminate and localize objects among background and other objects.
YOLOv3 \cite{YOLOv3} maintains detailed grid features to retain  detection accuracy of small objects. However, the effectiveness is limited, as accurate detection of both small and large objects cannot be kept together at the same time. The best performing method from LPIRC 2019 challenge \cite{2019LPIRC} shows improvement on detecting general-sized objects but not small objects on the COCO dataset \cite{COCO:ECCV:2014}. 

How to design fast and accurate network that can effectively detect all object sizes is still an open question. One solution to retain accurate feature localization is to add a bottom-up pathway to offset the lost information from low-level feature maps. In \cite{sermanet2013overfeat}, the adding of a gating module on the SSD frame leads to a gated bi-directional FP; however such gated network is not easily trainable. In \cite{PANet:CVPR:2018}, a bottom-up path aggregation network was proposed for object segmentation.  
The bi-directional network of \cite{LRF_Wang_2019_ICCV} can  efficiently circulate both low-level and high-level semantic information for small object detection. In \cite{EfficientDet:CVPR:2020}, a BiFPN was proposed based on NAS-FPN \cite{NAS-FPN:CVPR:2019} to better detect small objects with high efficiency.   In YOLOv4 \cite{YOLOv4:arXiv:2020}, path aggregation \cite{PANet:CVPR:2018} was modified by replacing the addition with concatenation for better detection of small objects. However, such BiFPN structure still cannot keep up accurate detections of both small and large objects all together.

\begin{figure*}[t]
\centerline{
  \includegraphics[width=0.9\textwidth]{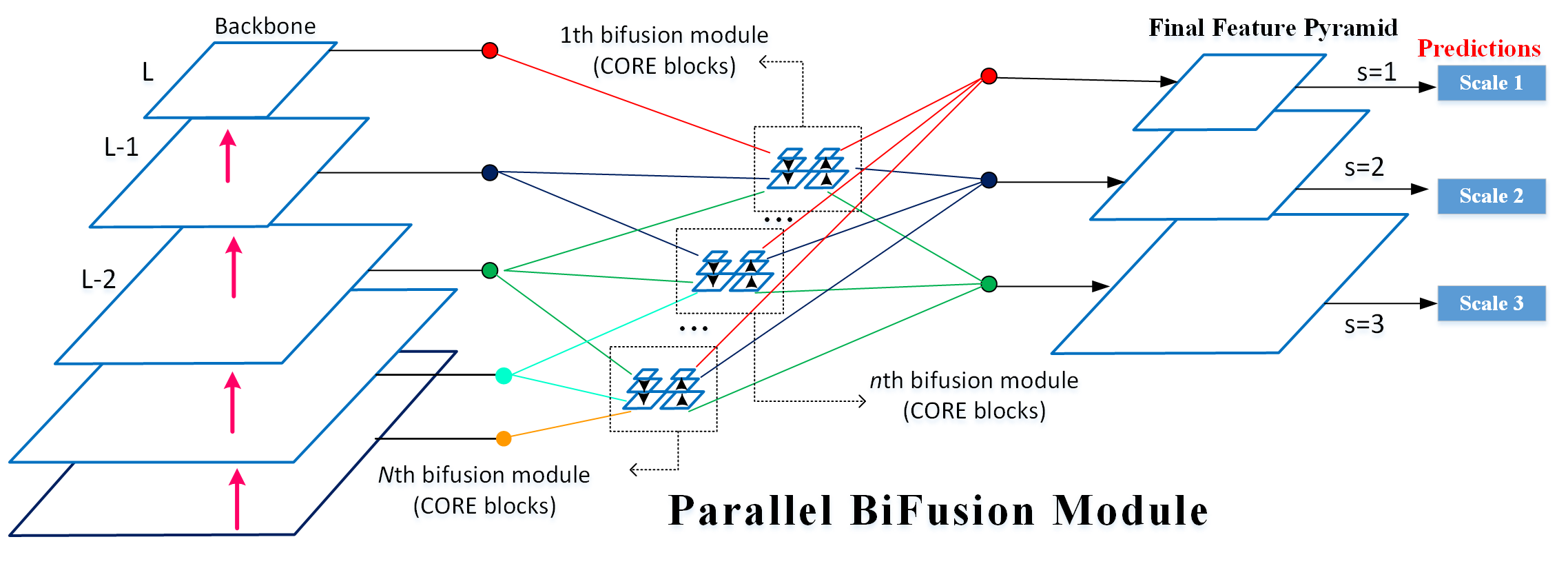} 
  \vspace{-0.3cm}
}
\caption{
Overview of the proposed {\bf Parallel Residual Bi-Fusion Feature Pyramid Network (PRB-FPN)}.
\vspace{-0.3cm}
}
\label{fig:overview}
\end{figure*}


We propose a new {\bf Parallel Residual Bi-Fusion Feature Pyramid Network (PRB-FPN)} with a parallel design and multiple improvements that can retain both deeper and shallower features for fast and accurate single-shot object detection.  Different from other bi-fusion FPN structures such as PANet \cite{PANet:CVPR:2018}, NAS-FPN \cite{NAS-FPN:CVPR:2019}, and BiFPN  \cite{EfficientDet:CVPR:2020}, we create a parallel bi-fusion structure to fuse three-layers of feature maps in parallel to generate three prediction maps at the same time, see Fig.~\ref{fig:overview}. Without losing efficiency, these three-way prediction maps can retain more accurate semantic and localization information to better detect both tiny and large objects. In this parallel structure, we introduce a new concatenation and re-organization (CORE) module for data fusion, where output features can be further purified to retain contextual information. We introduce a ``residual'' design (motivated from the spirit of ResNet~\cite{ResNet}) into our bi-fusion pipeline, which  enables easy training and integration with a number of popular backbones. We will show that our residual FP design outperforms other bi-directional methods \cite{LRF_Wang_2019_ICCV, Woo2019GatedBF} in Section~\ref{sec:results}. 
In comparison, methods based on traditional FPs \cite{FPN,YOLOv3,SSD,YOLOv2} can only learn un-referenced features, thus they are not suitable for detecting both large and small objects. Our residual FP retains semantic richer features in higher layers that can better detect small objects.

A key novelty in our design is the adding of {\em parallelization} to the bi-fusion FPN architecture. This parallel design is more effective in feature representation, {\em i.e.} for capturing features to identify and localize objects in either small or large sizes without losing efficiency.
In comparison, most existing bi-directional FP methods \cite{LRF_Wang_2019_ICCV,PFPNet:ECCV:2018,Woo2019GatedBF,Wu2018SingleShotBP} directly concatenate large feature maps in a memory-consuming way, which ends up with an even larger feature map. 

The proposed PRB-FPN is simple, efficient, and suitable for generic object detection for multiple object classes and sizes (small, mid, and large). We will show in Section~\ref{sec:results} that our approach is  generalizable in combining with mainstream backbones including Pelee~\cite{wang2018pelee} and DarkNet53~\cite{YOLOv3}. It can run in real-time and is easily deployable to edge devices. Main contributions of this paper are summarized in the following:

\begin{itemize} 
\item We propose a new Parallel Residual Bi-Fusion Feature Pyramid Network (PRB-FPN) that can effectively fuse both deep and shallow feature layers in parallel for fast and accurate one-shot object detection.

\item The parallel design of PRB-FPN makes it well-suited for detecting objects in both small and large sizes with higher accuracy.  
 
\item The PRB-FPN can be easily trained and integrated with different backbone thanks to the residual design. A newly proposed bottom-up fusion module (BFM) can improve the detection accuracy of both small and large objects.
        
\item Extensive experiments on Pascal VOC \cite{VOC} and MS COCO \cite{COCO:ECCV:2014} datasets show that PRB-FPN achieves the SoTA results for accurate and efficient object detection. Results also show great generalization ability on various object sizes and types.
    
\end{itemize}
    
\begin{figure*}[t]
\centerline{
  {\footnotesize (a)}
  \includegraphics[width=0.8\textwidth]{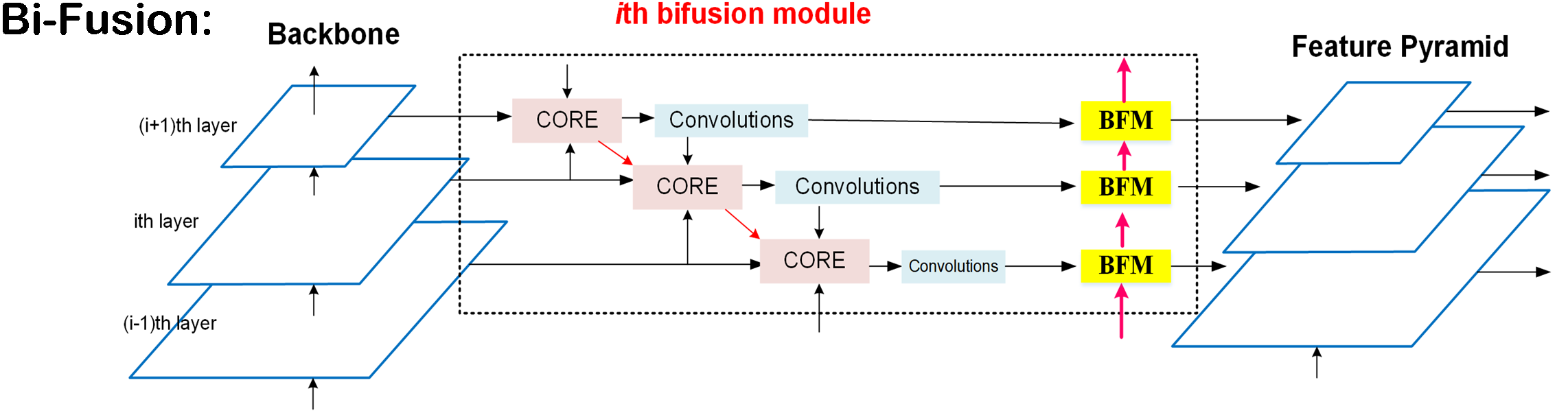}  
  \vspace{0.2cm}
}
\centerline{
  {\footnotesize (b)}
  \includegraphics[height=0.27\textwidth]{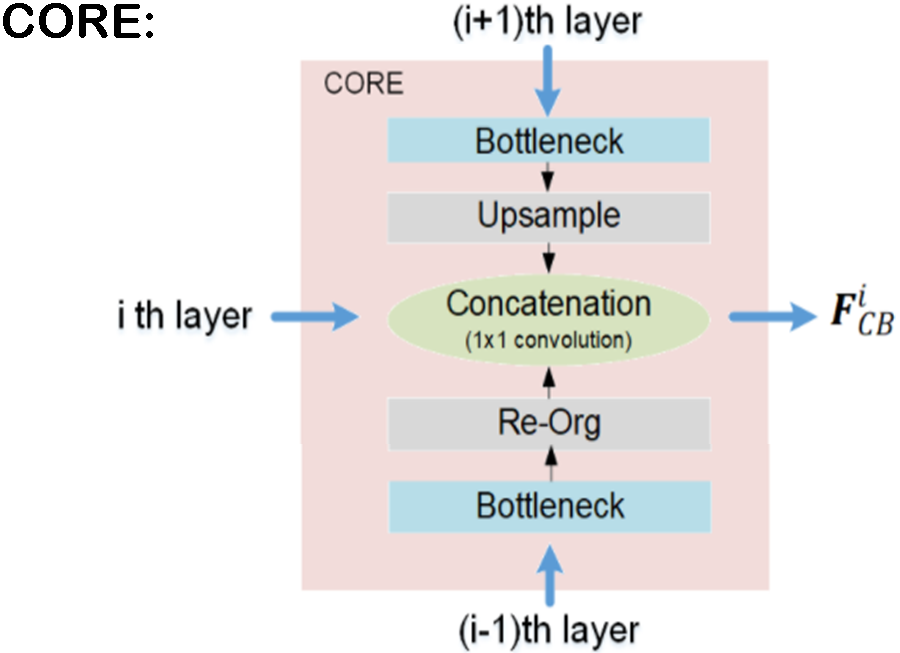}
  {\footnotesize (c)}
  \includegraphics[height=0.25\textwidth]{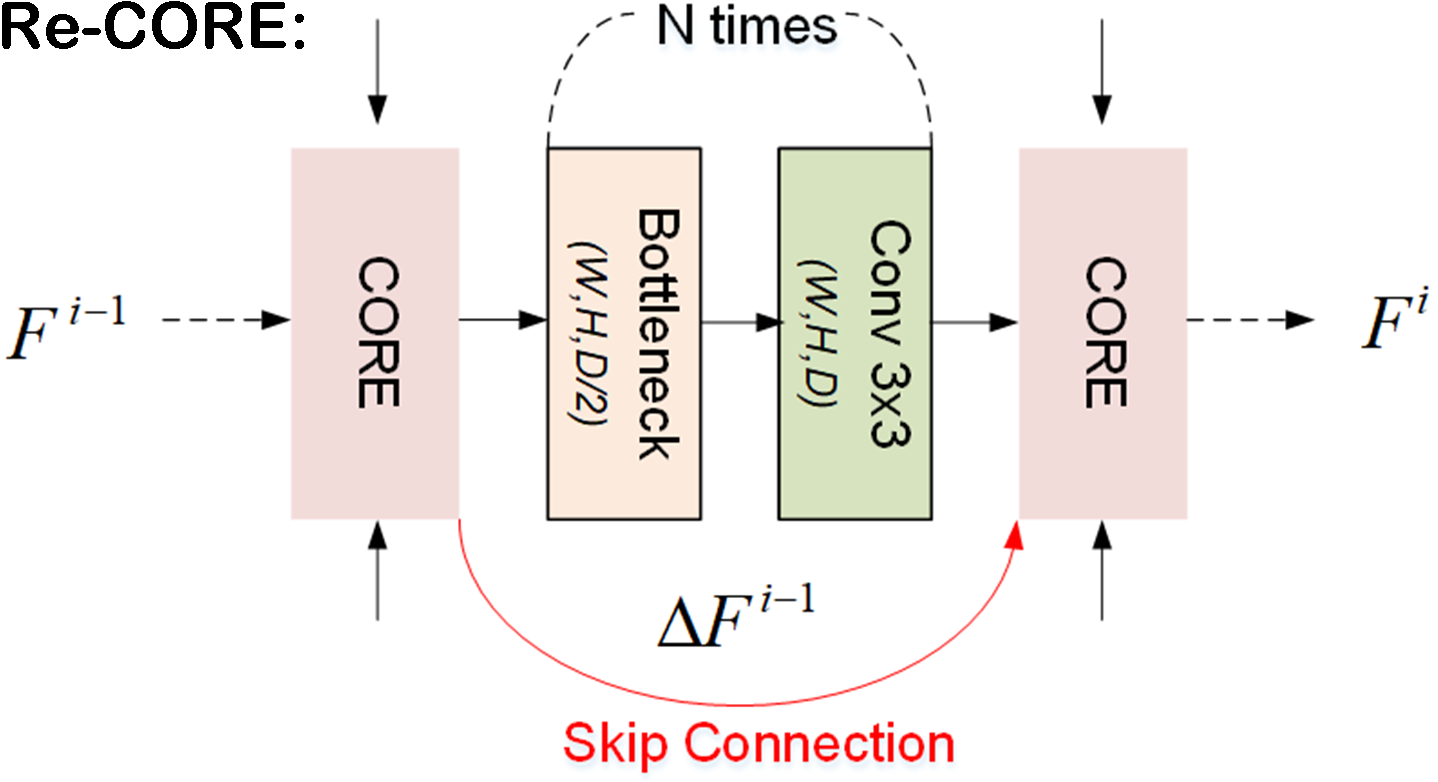}
  \vspace{-0.1cm}
  
}
\caption{
Detailed network architecture for the proposed modules (refer to Fig.~\ref{fig:overview} for the overall architecture of PRB-FPN): 
(a) the {\bf Bi-Fusion} module for concurrent fusion of contextual features from adjacent layers, 
(b) the {\bf concatenation and re-organization (CORE)} module for recursively fusion of contextual features from adjacent layers, and
(c) the {\bf Residual CORE (Re-CORE)} design in combining CORE with the residual design inspired from ResNet.
\vspace{-0.4cm}
}
\label{fig:Core:Puri}
\end{figure*}

\section{Related Works}
\label{sec:related}

{\bf Object detection} is a very active field in computer vision since the blooming of deep learning. The extensive amount of literature can be organized into two categories based on their network architectures: {\em two-stage} proposal-driven and {\em one-stage} (single-shot) approaches. In general, two-stage methods can achieve high detection accuracy but with longer computation time, while one-stage methods run faster with inferior accuracy. We focus on the survey of one-stage object detectors.
RefineDet \cite{RefineDet:CVPR:2018} employs an encode-decode structure in the deeper network with the use of up-sampling deeper scale features to enrich contextual information. PeLee~\cite{wang2018pelee} is a variant of DenseNet~\cite{huang2017densely} that outperforms SSD+MobileNet by 6.53\% on the Stanford Dogs dataset~\cite{StanfordDog} based on a much shallower network. However, PeLee \cite{wang2018pelee} does not detect small objects well on MS COCO~\cite{COCO:ECCV:2014}. PFPN~\cite{PFPNet:ECCV:2018} adopts the VGGNet-16 backbone~\cite{VGG:ICLR:2015} and SPP to generate a feature pyramid by concatenating multi-scale features. 


{\bf One-stage object detectors} mostly consist of a backbone network and a predictor. The backbone is a stacked feature map representing the input image in high feature resolution (but low spatial resolution for abstraction). The backbone network can be pre-trained as an image classifier on a large dataset such as ImageNet. OverFeat \cite{sermanet2013overfeat} was the first CNN-based one-stage object detector developed in 2013 with a sliding-window paradigm. Two years later, the first version of YOLO \cite{redmon2015unified} achieved state-of-the-art performance by integrating bounding box proposals and subsequent feature re-sampling in a single stage. SSD \cite{SSD} employed in-network multiple feature maps for detecting objects with varying shapes and sizes. The multi-map design enabled SSD with better robustness over YOLOv1 \cite{redmon2015unified}. For better detection of small objects, The Feature Pyramid Network (FPN) \cite{FPN} based on FP can achieve higher detection accuracy for small objects. YOLOv3 \cite{YOLOv3} was developed by adopting the concept of FPN.  By changing the backbone from DarkNet-19 \cite{YOLOv2} to DarkNet-53, YOLOv3 achieves superior performance in 2018.  Similarly, RetinaNet~\cite{Retina} combines FPN \cite{FPN} and ResNet \cite{ResNet} as the backbone. RetinaNet used focal loss to significantly reduce false positives in a single stage, such that the weights of each anchor box can be dynamically adjusted. 
Shift-invariance in CNNs was originally achieved using sub-sampling layers. The work of \cite{conf/icann/SchererMB10} evaluated the effect of small geometry perturbations on CNN and suggested that max-pooling is more effective in object detection and classification. 
In \cite{zhang2019shiftinvar}, a pooling-after-blurred technique was proposed by combing blurring and sub-sampling techniques to ensure shift-invariance. 

{\bf Feature pyramid (FP)} is widely used in SoTA detectors for detecting objects at different scales, where spatial and contextual features are extracted from the last layer of the top-down path for accurate object detection. This top-down aggregation is now a common practice for improving scale invariance in both two-stage and one-stage detectors. Popular FPs used for this purpose include the pyramidal feature hierarchy (bottom-up), hourglass (bottom-up and top-down), FPN~\cite{FPN}, SPP~\cite{SPP}, and PFPN~\cite{PFPNet:ECCV:2018}. It is also well-known that the top-down pathway in FP cannot preserve accurate object localization due to the shift-effect of pooling. 

{\bf Bi-directional FP} can recover lost information from shallow layers to improve {\bf small object detection} in several works \cite{LRF_Wang_2019_ICCV,Woo2019GatedBF,Wu2018SingleShotBP}. A gating module was used to control the feature flow direction in \cite{Woo2019GatedBF}. A light-weight scratch network and a bi-directional network were constructed in \cite{LRF_Wang_2019_ICCV} to efficiently circulate both low- and high-level semantic information. M2Det \cite{M2Det:AAAI:2019} is a one-stage detector that outperforms most 2019 methods on all multi-scale categories on MS COCO \cite{COCO:ECCV:2014}. However, the M2Det model is complicated and time-consuming, thus is not suitable for real-time object detection. Inspirited by NAS-FPN \cite{NAS-FPN:CVPR:2019}, a BiFPN was proposed in \cite{EfficientDet:CVPR:2020} to better detect small objects with higher efficiency. The recent YOLOv4 \cite{YOLOv4:arXiv:2020} modified the path aggregation method \cite{PANet:CVPR:2018} by replacing the addition with concatenation to better detect small objects. However, this BiFPN structure still cannot keep up accurate detection of both small and large objects all together.

{\bf Multi-Scale Object Detectors} face the challenge of small-size false positives due to the inadequacy of low-level features, which result in small receptive field size and weak semantic capabilities. The work of \cite{PreviewerFM:ACM:2018} demonstrates that independent predictions from different feature layers on the same region are beneficial in reducing false positives. In \cite{HGDNs:CVPR:2016}, a novel paradigm of multi-scale deep network is developed to model the spatial contexts surrounding different pixels at various scales. In \cite{MultiDefConv:2020:JourHCIS}, deep convolutional networks are used to obtain multi-scaled features, where deformable convolutional structures are added to overcome geometric transformations. 


{\bf Anchor-free methods} \cite{Law:2018:CDO,Wu2018SingleShotBP,tian2019fcos,AB:FSAF:CVPR:2019} do not reply on handcrafted anchors, thus are free of issues commonly associated with anchor-based designs. In \cite{Law:2018:CDO}, corner features are detected for object detection. By inheriting the architecture of R-CNN, ME R-CNN \cite{MERCNN:TIP:2020}  used multiple stream pipelines for accurate anchor-free object detection, where one pipeline is an expert for processing a certain type of ROIs and controlled by an expert assignment network.  A cascade anchor refinement module is proposed in \cite{Wu2018SingleShotBP} to refine pre-designed anchors. This is then injected into a bidirectional FP, which can detect objects with highly accurate localization. However, one pass of regression during training is not accurate enough for detection in this anchor-free approach. In \cite{AttentionCoupleNet:TIP:2019}, an attention CoupleNet was proposed by designing a cascade attention structure to generate class-agnostic attention maps of target regions so that a discriminative feature representation can be formulated for part-based object detection.  In  \cite{SAFNet:TIP:2020}, Jin {\em et al.} used an adaptive anchor generator to generate all possible anchor boxes. They then proposed a semi-anchor-free network for object detection with an enhanced feature pyramid which consists of two modules, {\em i.e.}, adaptive feature fusion module (AFFM) and self-enhanced module (SEM).  In \cite{tian2019fcos}, a number of low-quality bounding boxes are predicted and further verified with a {\em centerness} branch that can detect objects without using any pre-defined anchor boxes. In \cite{Cao_2019_ICCV}, a hierarchical shot detector is used to predict detection bounding boxes via regression. These regression based methods are more accurate but less efficient.  Compared with CornerNet \cite{Law:2018:CDO}, FoveaBox \cite{FoveaBox:TIP:2020} does not require any embedding or grouping techniques at post-processing stage to locate real bounding boxes.  However, its latency is higher and results in lower efficiency.


{\bf Network Transferring} The above detectors can be trained well enough from a large set of sufficiently representative data.  However, as there exists numerous application scenarios in which only a few training samples ({\em e.g.} tumor images in medical applications) are available, transfer learning can be used to customize the model and adopt to the tasks~\cite{TL:ITNN:2015}. For example, in \cite{Weakly-Shared:ACMM:2015}, a weakly-shared Deep Transfer Network (DTN) was proposed to hierarchically learn and transfer semantic knowledge from web texts to images for image classification.  In \cite{Generalized-Deep:ACMM:2016}, a novel generalized DTNs was proposed to solve the problem of insufficient training images by transferring label information across heterogeneous domains, such as transferring from the textual to visual domain for image classification.  Moreover, in  \cite{GAIA}, a transfer learning system named GAIA was proposed to provide powerful pre-trained weights, select models, and collect relevant data for object detection when only a few training samples were given. However, although network transferring methods can provide performance improvements, they cannot outperform ordinary methods trained with sufficient samples.


\section{Method}
\label{sec:method}

\begin{figure*}[t]
\centerline{
\includegraphics[width=\textwidth]{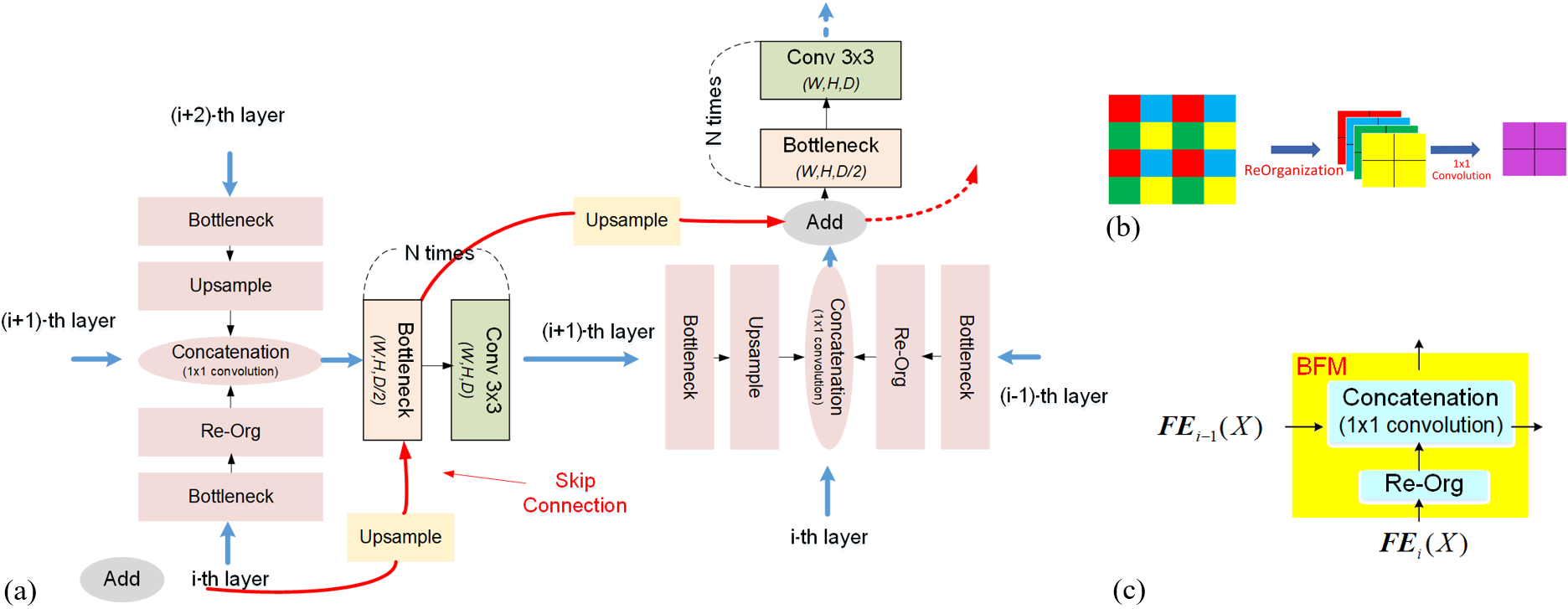}
}
\caption{
(a) Details of the proposed {\bf Re-CORE} architecture.
(b) The {\bf Re-Org} block for feature re-organization.
(c) The {\bf bottom-up fusion module (BFM)}.
\vspace{-0.3cm}
}
\label{fig:CORE:BFM}
\end{figure*}

We first motivate the design of our proposed network architecture by addressing the limitation of the Feature Pyramid (FP) for visual object detection.
In Section~\ref{sec:ParallelBiFusion}, details of our new parallel bi-fusion scheme are described.  The adding of {\em parallelization} to the bi-fusion FPN architecture can better capture features for both small and large objects without degrading efficiency. 
Section~\ref{sec:CORE} describes our new feature concatenation and re-organization scheme that can effectively circulate semantic and localization information.
Section~\ref{sec:Res:BiFusion:FP} further adopts a {\em residual recursive} formulation into our pipeline, which enables easier training and better performance for small object detection.
Finally, Section~\ref{sec:BFM} adds one more design of bottom-up location feature fusion that can further improve object localization. Figs. \ref{fig:overview}, \ref{fig:Core:Puri} and \ref{fig:CORE:BFM} depict the complete pipeline of our proposed network architecture. Details are provided in the following sessions.


\subsection{Parallel Concatenation and Re-organization Feature Bi-Fusion Architecture}
\label{sec:ParallelBiFusion}

Feature Pyramid (FP) is widely-used in top-down feature aggregation that can collect semantically rich features to effectively discriminate objects with scale invariance. However, it is well-known that FP cannot preserve accurate localization for small objects due to pooling and quantization. The winning methods of LPIRC 2019 challenge \cite{2019LPIRC} show improvements on detecting general-sized objects but not on small objects. There object prediction was carried out using information from both each pyramid layer and the respective lower layers. This coincide with thoughts from several SoTA bi-directional methods \cite{conf/icann/SchererMB10,PFPNet:ECCV:2018,Woo2019GatedBF} in leveraging new feature streams from lower feature layers (or the raw image itself) to keep track of features from smaller objects and achieve more accurate localization. Such bi-fusion modules specially designed for improving small object detection still lack capabilities in detecting larger objects. 

In this paper, we propose an effective {\em parallel} FP fusion design to tackle this difficult problem of {\em object detection considering all object scales}. This is done by creating multiple bi-fusion paths to keep tracks of features that are suitable to detect objects of all sizes (including tiny and large objects). Each bi-fusion path keeps track of size-dependent features to represent objects at a specific scale. Assume that there are $N$ prediction maps (where $N=3$ for YOLOv4), we propose to execute the $N$ different concurrent fusion paths to generate $N$ fused feature maps for the $N$ prediction maps. We use $n$ to index the bi-fusion modules (thus $n\leq N$). Let $L$ be the level of the top layer in the FP. As shown in Fig.~\ref{fig:overview}, the $n^{th}$ bi-fusion module will {\em bi-fuse} feature maps from the $(L-n+1)^{th}$ layer to the $(L-n-N+2)^{th}$ layer in the backbone.  The $s^{th}$ output will be fed into the $s^{th}$ prediction map for object detection,
which will integrate feature maps from the $s^{th}$ layer of all bi-fusion modules.  Noticeably, the 
$1^{st}$
bi-fusion module in our model corresponds to the sole bi-fusion module in SoTA bi-directional methods \cite{conf/icann/SchererMB10,PFPNet:ECCV:2018,Woo2019GatedBF}.



\subsection{Concatenation and Re-organization for Feature Bi-Fusion}
\label{sec:CORE}

In Fig.~\ref{fig:Core:Puri}, each bi-fusion module consists of three {\bf concatenation and re-organization (CORE)} blocks and two skip connections. Details of the bi-fusion module are shown in Fig.~\ref{fig:Core:Puri}(a).   The {\bf CORE} design in Fig.~\ref{fig:Core:Puri}(b) brings a major advantage that feature fusion can be recursively applied in both top-down and bottom-up fashions to: 
(1) concatenate semantic features from top layers (top-down), and 
(2) re-organize spatially rich localization features from bottom layers (bottom-up). 
To avoid using too many dithering operations ({\em i.e.}, point-wise convolutions) and to avoid computationally expensive operations ({\em i.e.}, pooling and addition), we adopt an $1 \times 1$ depth-wise convolution in the CORE module. This enables effective fusion of pathways coming from deeper and shallower layers in each layer of the FP. Our $1 \times 1$ depth-wise convolution in CORE is very different from most of SoTA bi-directional methods \cite{conf/icann/SchererMB10,PFPNet:ECCV:2018,Woo2019GatedBF,Wu2018SingleShotBP}, where feature fusion is carried out by concatenating all feature maps. Their simple concatenations result in a large feature map proportional to the total feature size. In contrast, our $1 \times 1$ conv filter in CORE is automatically learned, such that features can be fused more effectively via a feature map of fixed size. 

In each layer of the used backbone, CORE fuses features of each layer with its two adjacent (immediately shallower and deeper) layers. In other words, feature {\em bi-fusion} is performed in the feature pyramid of CORE. In the bottom-up fusion with the shallower layer, similar to YOLOv2~\cite{YOLOv2}, a {\bf Re-Org} block from Fig.~\ref{fig:CORE:BFM}(b) is adopted in Fig.~\ref{fig:Core:Puri}(b) to re-organize the feature map into 4 channels. However, instead of using a concatenation operation, the $1 \times 1$ convolution filter is then performed to fuse all feature maps as the output.

\subsection{Residual Bi-Fusion Feature Pyramid}
\label{sec:Res:BiFusion:FP}


We further adopt the residual concept inspired from ResNet~\cite{ResNet} to the CORE block in our design, and created a new {\bf Residual CORE (Re-CORE)} block. Re-CORE enables the fusion of four adjacent scales  (namely, the {\em shallow}, {\em current}, {\em deep}, and {\em deeper} layers) for better detection of small objects.
Specifically, by recursively injecting the output of the $(i+1)^{th}$ CORE module to the $i^{th}$ CORE module, the Bi-Fusion FP becomes a fully-featured {\bf Residual Bi-Fusion FPN} as in Fig.~\ref{fig:CORE:BFM}(a). 
Fig.~\ref{fig:Core:Puri}(c) depicts the connection between the Re-CORE and Convolution modules, in which $F^i$ and  $\Delta F^i$  denote the outputs of  the $i$-th Re-CORE and Convolution modules, respectively.

Fig.~\ref{fig:CORE:BFM}(a) shows the detailed Re-CORE architecture. The Re-CORE module performs bi-fusion to integrate features from the four input layers with residual design. The output of the previous Re-CORE module becomes the input of the current Re-CORE module via a {\em skip connection}, which is depicted as a red line in Fig.~\ref{fig:Core:Puri}(c) and Fig.~\ref{fig:CORE:BFM}(a), respectively.  
Features from the $i^{th}$, $(i-1)^{th}$, and $(i+1)^{th}$ layers are fused by an $1\times1$ convolution and then added to an up-sampled version of the skip connection to produce a new feature map. This map is then fed into a convolution block to produce the final output of this Re-CORE module.  

{\bf Working with popular backbones:}
Similar to ResNet \cite{ResNet}, the residual nature of our Re-CORE module enables easy training and integration of the FP with a wide range of backbones that works particularly well for small object detection. Instead of learning un-referenced features, Re-CORE obtains better accuracy from the largely increased feature depths when compared with traditional FPs \cite{FPN,YOLOv3,YOLOv2,SSD}. Note that SoTA FPs \cite{FPN,SPP,PFPNet:ECCV:2018} often learn redundant features and perform poorly on small object detection.  


\begin{figure}
\centerline{
  \includegraphics[width=0.75\textwidth]{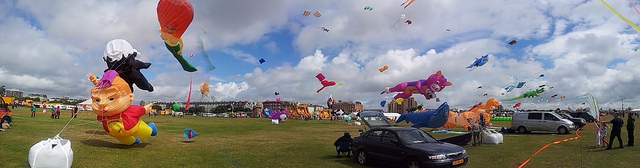}
\vspace{-0.15cm}
}
\centerline{
  {\footnotesize (a) an image fron the COCO-test-dev}
\vspace{0.15cm}
}  
\centerline{
  \includegraphics[width=0.75\textwidth]{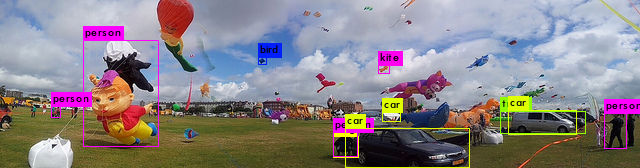}
\vspace{-0.15cm}
}
\centerline{
  {\footnotesize (b) YOLOv3 512 $ \times $ 512}
\vspace{0.15cm}
}  
\centerline{
  \includegraphics[width=0.75\textwidth]{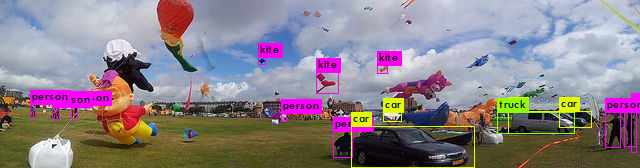}
\vspace{-0.15cm}
}
\centerline{
  {\footnotesize (c) YOLOv3 with BFM 512 $ \times $ 512}
\vspace{0.15cm}
}  
\centerline{
  \includegraphics[width=0.75\textwidth]{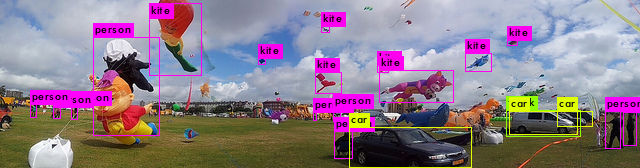}
\vspace{-0.15cm}
}
\centerline{
  {\footnotesize (d) YOLOv3 with Re-CORE 512 $ \times $ 512}
\vspace{0.15cm}
}  
\centerline{
  \includegraphics[width=0.75\textwidth]{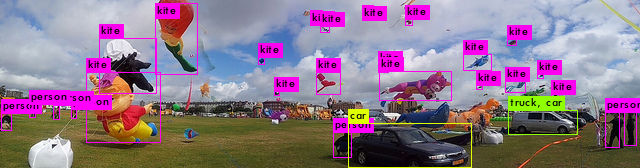}
\vspace{-0.15cm}
}
\centerline{
  {\footnotesize (e) PRB-FPN 512 $ \times $ 512}
\vspace{0.15cm}
}  
\centerline{
  \includegraphics[width=0.75\textwidth]{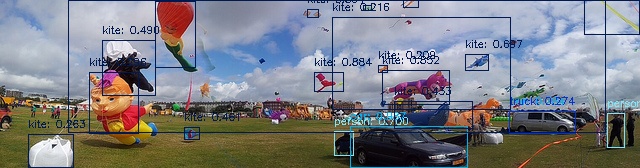}
\vspace{-0.15cm}
}
\centerline{
  {\footnotesize (f) M2Det~\cite{M2Det:AAAI:2019} 512 $ \times $ 512}
\vspace{-0.2cm}
}  
\caption{ 
Small object detection results on the MS COCO test set.
}
\label{fig:ablationstudyvisulize}
\end{figure}

The Re-CORE module provides a new effective fusion approach for collecting localization information from bottom layers that can improve the accuracy of small object detection. 
In comparison, the naive approach in \cite{FindTinyFaces2017} detects small objects by generating high-resolution images as inputs to the detection module, which comes with a cost of large computational burden. 
Another approach for small object detection is to leverage contextual information, by sending semantic features from a top-down way via a FP as in YOLOv3 \cite{YOLOv3}. However, in these methods without the use of residual property, the learning will include un-referenced features and thus bound the number of FP layers that can actually contribute to object detection.  In comparison to the FP proposed in \cite{2019LPIRC}, our Re-CORE module can capture richer semantic features from deeper layers that can directly improve small object detection. 

In summary, our residual design and bi-directional fusion make the Re-CORE module suitable for detecting small and  even tiny objects without notable computation overheads. 

\subsection{Bottom-up Feature Fusion}
\label{sec:BFM}

As aforementioned, the top winning method in LPIRC 2019 \cite{2019LPIRC} improved detection on large and medium-sized objects, but not able to keep up the performance for small objects. 
To address this issue, we propose the adding of a {\bf bottom-up fusion module (BFM)} to the PRB-FPN network to  further improve the localization of both small and large objects. Fig.~\ref{fig:CORE:BFM}(c) depicts the proposed BFM architecture. Instead of using convolution with stride 2 (adopted in PANet \cite{PANet:CVPR:2018}, Bi-FPN \cite{EfficientDet:CVPR:2020}, or YOLOv4 \cite{YOLOv4:arXiv:2020}), the BFM adopts a Re-Org block to split $C$ channels of feature map into $4C$ channels to better preserve spatial information and generate robust semantic features via $1 \times 1$ convolution, which improves small object detection.  As for the bidirectional FPN-BPN work~\cite{Wu2018SingleShotBP}, convolutions with stride 2 are used for down-sampling, while de-convolutions are adopted for up-sampling. However, this design results in lower accuracy for small object detection due to the stride 2 operator, and the use of de-convolution leads to low efficiency in object detection.

In summary, in our design: (1) Section~\ref{sec:ParallelBiFusion} describes the use of parallel bi-fusion paths that run concurrently for effective detection of both small and large objects; (2) Section~\ref{sec:Res:BiFusion:FP} describes the adding of the Re-CORE module for improving the detection of small objects; and finally, (3) the BFM in Section~\ref{sec:BFM} can bring specific local information from a bottom-up pathway to localize the objects more accurately. The BFM pathway works particularly well for detecting both large and mid-sized objects. Experimental results in this regard are shown in Table~\ref{tbl:ReCORE:BFM}.



\section{Experimental Results}
\label{sec:results}

We evaluate the PRB-FPN against SoTA object detection methods on MS COCO benchmark~\cite{COCO:ECCV:2014} and UAVDT \cite{UAVDT} using machines with nVidia Titan X GPU and V100. Accuracy is evaluated in the metric of Average Precision (AP). Computational efficiency is evaluated in the processing frames per second (FPS).

{\bf Backbones.} Our pipeline is not limited to any feature extraction backbone. We evaluated the following backbones: PeLee~\cite{wang2018pelee}, MobileNet-V2~\cite{MobileNet-V2}, DarkNet-53~\cite{YOLOv3}, VGG16~\cite{VGG:ICLR:2015}, \texttt{ResNet-50}~\cite{ResNet}, DenseNet~\cite{huang2017densely}, CSPnet~\cite{CSPNet}.

\subsection{Implementation Details and Evaluation Configures}
\label{subsec:impDetail}

{\bf Implementation details:} For performance evaluations on MS COCO dataset, the default hyper-parameters are set as follows. Total training steps are $500,500$ with the step decay learning rate 0.001. The learning rate is further multiplied by a factor $0.01$ at the $400,000$ steps and $450,000$ steps, respectively. Momentum and weight decay rate are set to be $0.9$ and $0.0005$, respectively. All various PRB models were trained on a single V100 with batch size $64$, and mini-batch size $16$, $8$, or $4$ depended on the used model size for fitting the limitation of the available GPU RAM.

{\bf Evaluation details:}  We next evaluate the newly introduced designs of PRB-FPN in terms of how each design effectively fuse both deep and shallow feature  layers in parallel for fast and accurate one-shot object detection. The first major design in the parallel structure of PRB-FPN is the new {\em residual concatenation and re-organization} (ReCORE) module proposed for effective and efficient data fusion. 
The second major design is the {\em bottom-up fusion module} (BFM) added after ReCORE in PRB-FPN as shown in Fig.~\ref{fig:Core:Puri}, which can further improve the localization of both small and large objects. To evaluate the effect of each module, accuracy improvement based on BFM is first evaluated in $\S$~\ref{sec:eval:BFM}. The effect of BFM with ReCORE module for accuracy improvements is evaluated in $\S$~\ref{sec:eval:ReCORE}. Evaluation of the RB-FPN module against the original RPN is provided in $\S$~\ref{sec:eval:Orig:FPN}. Finally, comparisons between  PRB-FPN and other SoTA methods are provided in $\S$~\ref{sec:eval:SOTA}.

\begin{table}[t]
\caption{Ablation study of BFM among different backbones.
  \vspace{-0.1cm}
}
\centerline{
    \setlength\tabcolsep{1.3pt}
    \scriptsize
    \begin{tabular}{ccccccccc}
    \toprule
     Backbone & BFM & FPS & AP & AP$_{50}$ & AP$_{75}$ & AP$_s$ & AP$_M$ & AP$_L$ \\
    \midrule
     DarkNet53   &  & \textbf{28.9} & 28.6 & 50.7 & 29.6 & 15.5 & 30.4 & 35.3 \\ 
      512x512  &  $\checkmark$  & 28.4 & \textbf{34.9} & \textbf{57.2} & \textbf{37.7} & \textbf{18.6} & \textbf{37.1} & \textbf{45.3} \\ 
    \midrule
    Pelee  & & \textbf{85.8} & 26.7 & 49.9 & 26.2 & 13.5 & 27.8 & 33.5 \\ 
      512x512 &  $\checkmark$  & 84.5 & \textbf{28.3} & \textbf{51.8} & \textbf{28.4} & \textbf{14.0} & \textbf{30.1} & \textbf{35.6} \\ 
    \midrule
    VGG16  & & \textbf{31.8} & 34.1 & 58.3 & 35.8 & 17.9 & 35.9 & 44.1 \\ 
      512x512 &  $\checkmark$  & 31.4 & \textbf{34.6} & \textbf{58.6} & \textbf{36.7} & \textbf{18.6} & \textbf{36.5} & \textbf{44.3} \\ 
    \midrule
    DenseNet201 & & \textbf{30.5} & 30.1 &54.5 &32.5 &15.7 &33.8 &40.8 \\
    512x512 &  $\checkmark$ &39.4 &\textbf{31.5} &\textbf{54.7} &\textbf{33.3} &\textbf{15.9} &\textbf{33.9} &\textbf{41.1} \\
    \bottomrule
    \end{tabular}
}
\vspace{-0.15cm}
\label{tbl:backbones}
\end{table}

\begin{table}[t]
\caption{Comparisons between our BFM and other SoTA bi-directional fusion methods.
  \vspace{-0.1cm}
}
\centerline{
    \setlength\tabcolsep{1.3pt}
    \scriptsize
    \begin{tabular}{cccccccccc}
    \toprule
     Method  & Backbone & Input size & FPS & AP & AP$_{50}$ & AP$_{75}$ & AP$_s$ & AP$_M$ & AP$_L$ \\
    \midrule
    GBFPN-SSD~\cite{Woo2019GatedBF} & VGG16 & 512$\times$512 & - & -  & 33 & -& -&-&- \\
    FPN-BPN~\cite{Wu2018SingleShotBP} & VGG16 & 320$\times$320 &\textbf{32.4} & 29.6 & 48.4  & 32.3 & 9.6 & 32.5 & 44.3\\
    FPN-BPN~\cite{Wu2018SingleShotBP} & VGG16 & 512$\times$512 &18.9 & 33.1 & 53.1  & 36.3 & 15.7 & \textbf{37} &44.2\\
    EfficientDet-D0~\cite{EfficientDet:CVPR:2020} & EfficientNet~\cite{EfficientNet:pmlr:2019} & 512$\times$512 &- & 34.6 & 53.0  & 37.1 & - & - &-\\
    NAS-FPN~\cite{NAS-FPN:CVPR:2019} & ResNet-50 & 1024$\times$1024 &- & 44.2 & -  & - & - & - &-\\
    \midrule  
    BFM [Ours] & VGG16 & 512$\times$512 &31.4 & 34.6 & 58.6 & 36.7 & 18.6   & 36.5 & 44.3 \\
    RB-FPN [Ours] & ResNet-50  & 512$\times$512  &32.1 & \textbf{44.3} & \textbf{65.1} & \textbf{48.2} & \textbf{25.1} & \textbf{47.3} & \textbf{56.8}\\
    \bottomrule
    \end{tabular}
}
\vspace{-0.15cm}
\label{tbl:biFPNCompare}
\end{table}

\begin{table}[t]
\centerline{
    \setlength\tabcolsep{0.8pt}
    \scriptsize
    \begin{tabular}{ccccccccccc}
    \toprule
    Method & Backbone & Re-CORE & BFM & FPS & AP & AP$_{50}$ & AP$_{75}$ & AP$_s$ & AP$_M$ & AP$_L$ \\
    \midrule
    Pelee$^*$ &Pelee$^*$  & &   &\textbf{85.8} &26.7 &49.9 &26.2 &13.5 &27.8 &33.5\\
    Pelee with BFM&Pelee &  & $\checkmark$ &84.5 &\textbf{28.3} &\textbf{51.8} &\textbf{28.4} &\textbf{14.0}&\textbf{30.1} &\textbf{35.6}\\
   Pelee with RB-FPN & Pelee & $\checkmark$ & $\checkmark$ &84.2 &\textbf{29.5} &\textbf{52.9} &\textbf{30.2} &\textbf{14.9}&\textbf{33.1} &\textbf{36.7}\\
    \midrule
    Yolov3$^\dagger$ &Darknet53$^\dagger$ & &    &\textbf{28.9} &32 &56.5 &33 &17.4 &34 &41.4 \\
    Yolov3-SPP$^\dagger$ &Darknet53$^\dagger$ & &    &28.7 &35.3 &59.2 &37.4 &16.9 &37.1 &48 \\
    Yolov3$^\ddagger$& Darknet53$^\ddagger$ & &    &\textbf{28.9} &28.6 &50.7 &29.6 &15.5 &30.4 &35.3\\
    Yolov3 with Re-CORE&Darknet53 &$\checkmark$ &  &27.6 &36 &59.5 &38.2 &18.9 &37.3 &47.1\\
    Yolov3 with BFM&Darknet53 &  &$\checkmark$ &28.4 &34.9 &57.2 &37.7 &18.6 &37.1 &45.3\\
    Yolov3 with RB-FPN & Darknet53 & $\checkmark$ & $\checkmark$ &\textbf{27.2} &\textbf{36.8} &\textbf{59.7} &\textbf{39.6} &\textbf{19} &\textbf{39.5} &\textbf{48}\\
     \midrule
    Yolov4 &CSPDarknet53 & &    &\textbf{31} &43 &64.9 &46.5 &24.3 &46.1 &55.2 \\
    Yolov4 with Re-CORE&CSPDarknet53 &$\checkmark$ &  &28.5 &44.8 &66.5 &47.3 &26.9 &46.3 &55.8\\
    Yolov4 with BFM&CSPDarknet53 &  &$\checkmark$ &30.5 &43.7 &65.3 &47.1 &24.5 &48.2 &55.3\\
    Yolov4 with RB-FPN & CSPDarknet53 & $\checkmark$ & $\checkmark$ &27.3 &\textbf{45.1} &\textbf{67.2} &\textbf{48.2} &\textbf{27.1} &\textbf{48.5} &\textbf{57}\\
   \bottomrule
  \end{tabular}
  \caption{
Ablation study of Re-CORE and BFM; RB denotes the proposed Residual Bi-Fusion design as in Fig.\ref{fig:CORE:BFM}(a).
\vspace{-0.1cm}
}
}  
\label{tbl:ReCORE:BFM}
\begin{flushleft}
\footnotesize
\hspace{0.2cm}$^*$  The input size for all backbones is 512x512. \\
\hspace{0.2cm}$^*$  Trained and tested by ourselves according to the paper.\\
\hspace{0.2cm}$^\dagger$ Test results with weights provided in the YOLOv3 website.\\
\hspace{0.1cm}$^\ddagger$ Trained and tested by ourselves according to the instruction.   
\vspace{-0.3cm}
\end{flushleft} 
\end{table}

\begin{figure*}[t]
\centerline{
  {\footnotesize (a)}
  \includegraphics[width=0.45\textwidth]{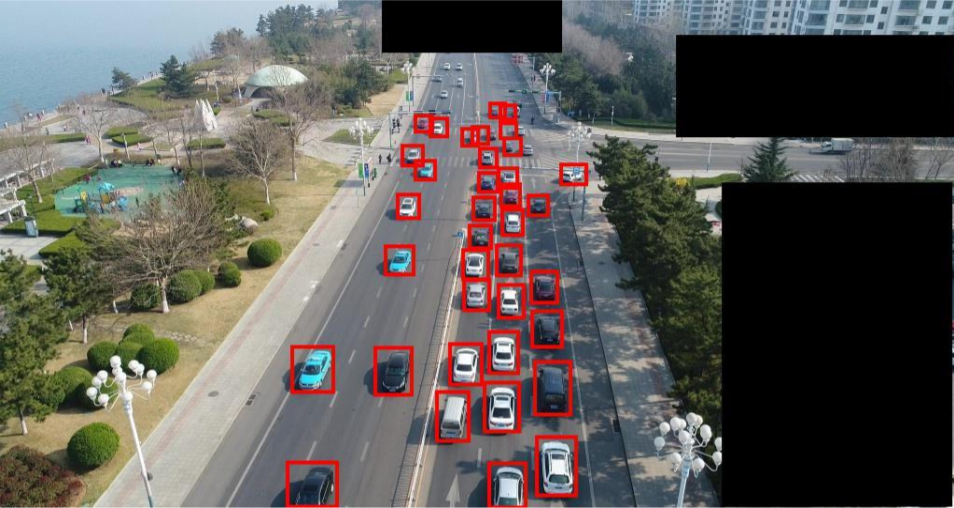}
  {\footnotesize (b)}
  \includegraphics[width=0.45\textwidth]{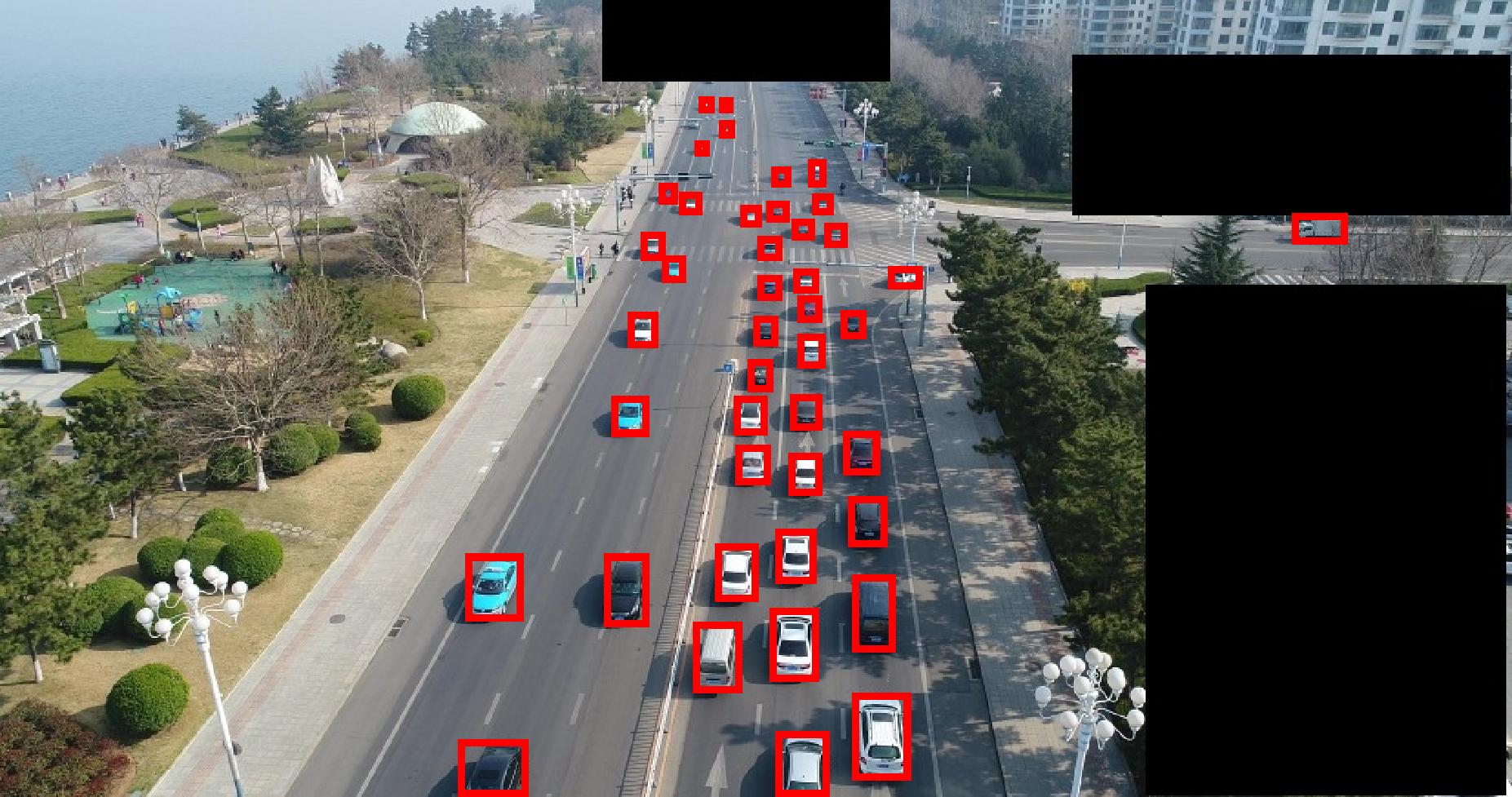}  
\vspace{-0.2cm}
}  

\centerline{
  {\footnotesize (c)}
  \includegraphics[width=0.45\textwidth]{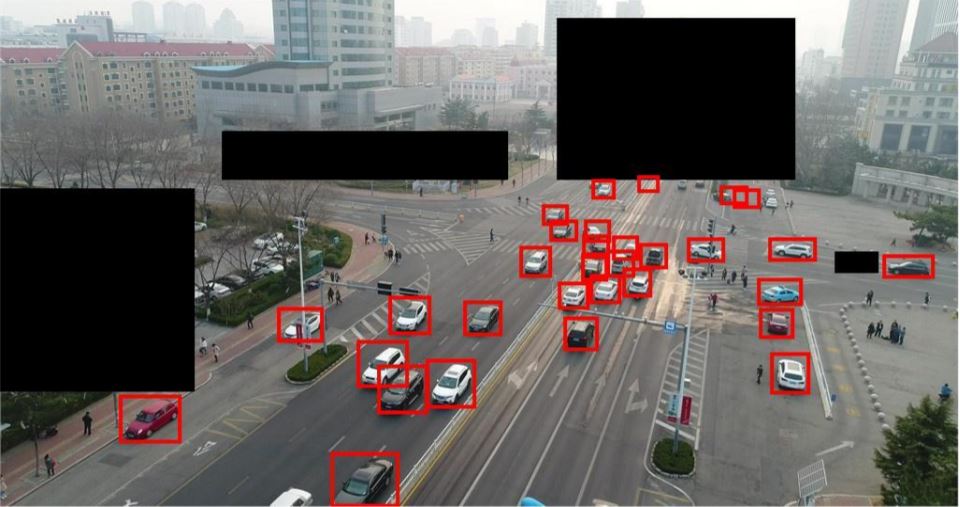}
  {\footnotesize (d)}
  \includegraphics[width=0.45\textwidth]{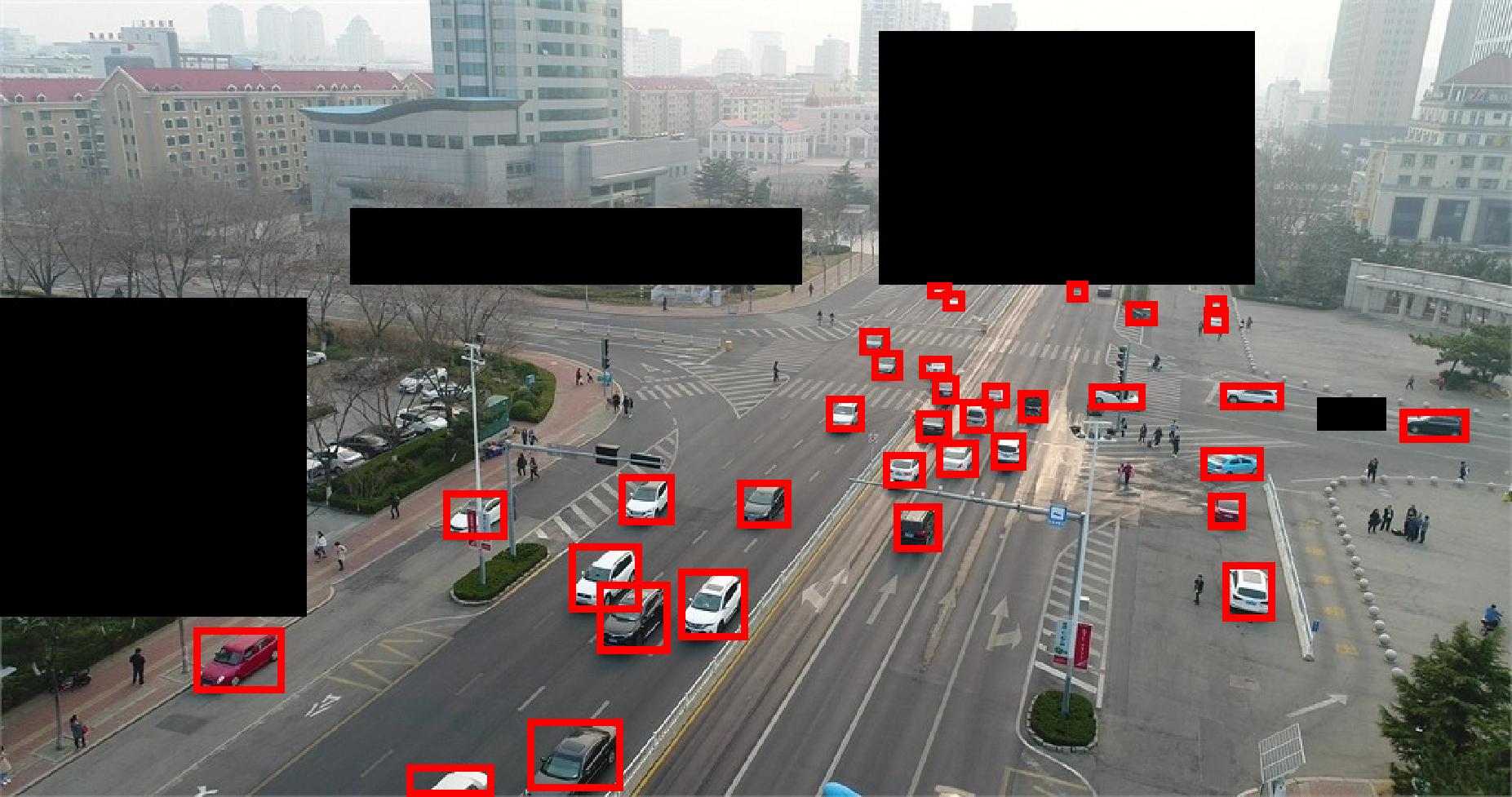}  
\vspace{-0.2cm}
}  

\caption{
Small object detection results on the UAVDT17 benchmark~\cite{UAVDT}.
(a) and (c): LRFNet~\cite{LRF_Wang_2019_ICCV}.
(b) and (d): The proposed PRB-FPN. Black boxes indicate don't-care regions that come with the original UAVDT17 dataset. 
\vspace{-0.3cm}
}
\label{fig:SOD}
\end{figure*}

\begin{figure*}[t]
\centerline{
  {\footnotesize (a)}
  \includegraphics[width=0.45\textwidth]{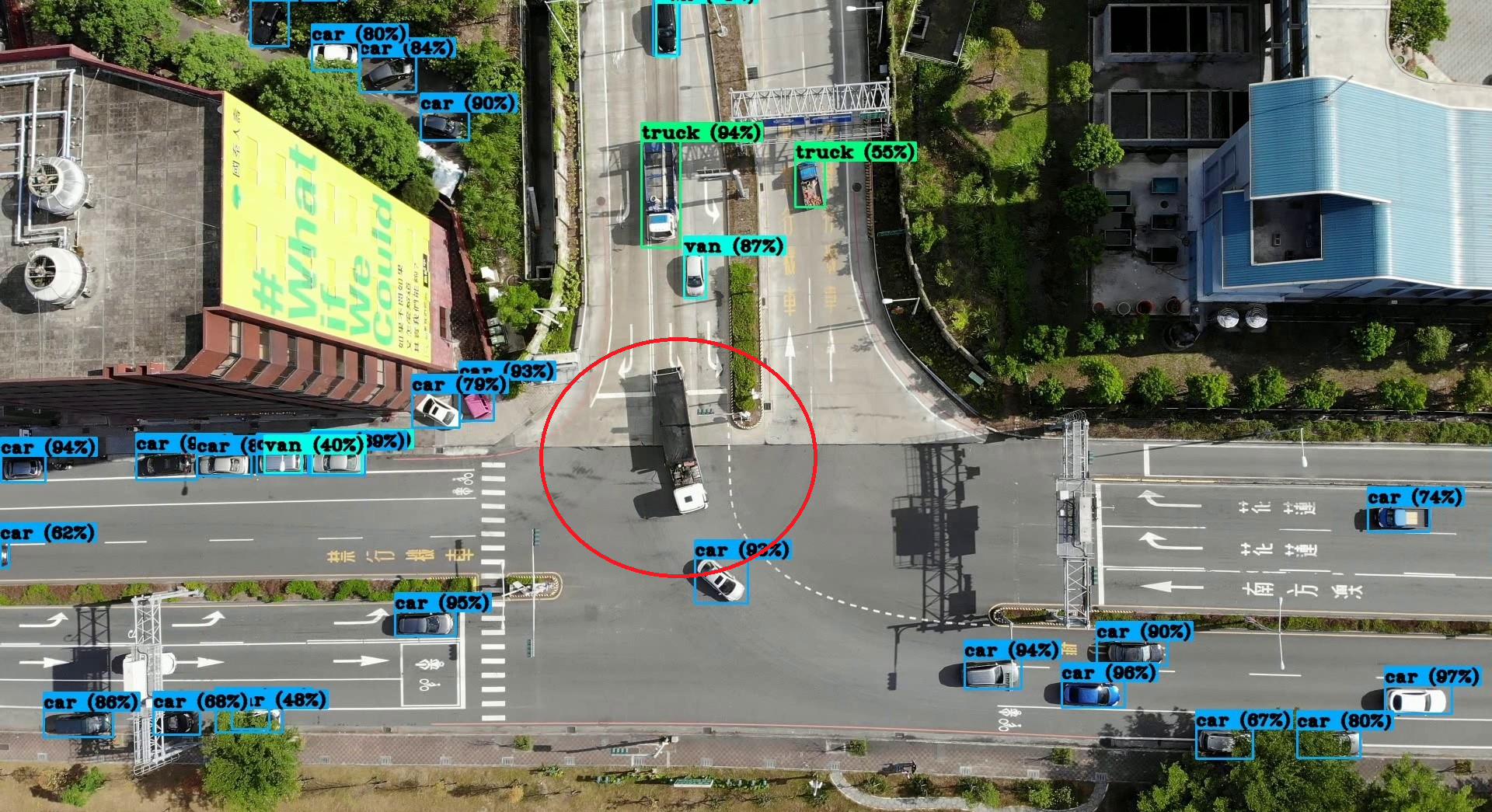}
   {\footnotesize (b)}
  \includegraphics[width=0.45\textwidth]{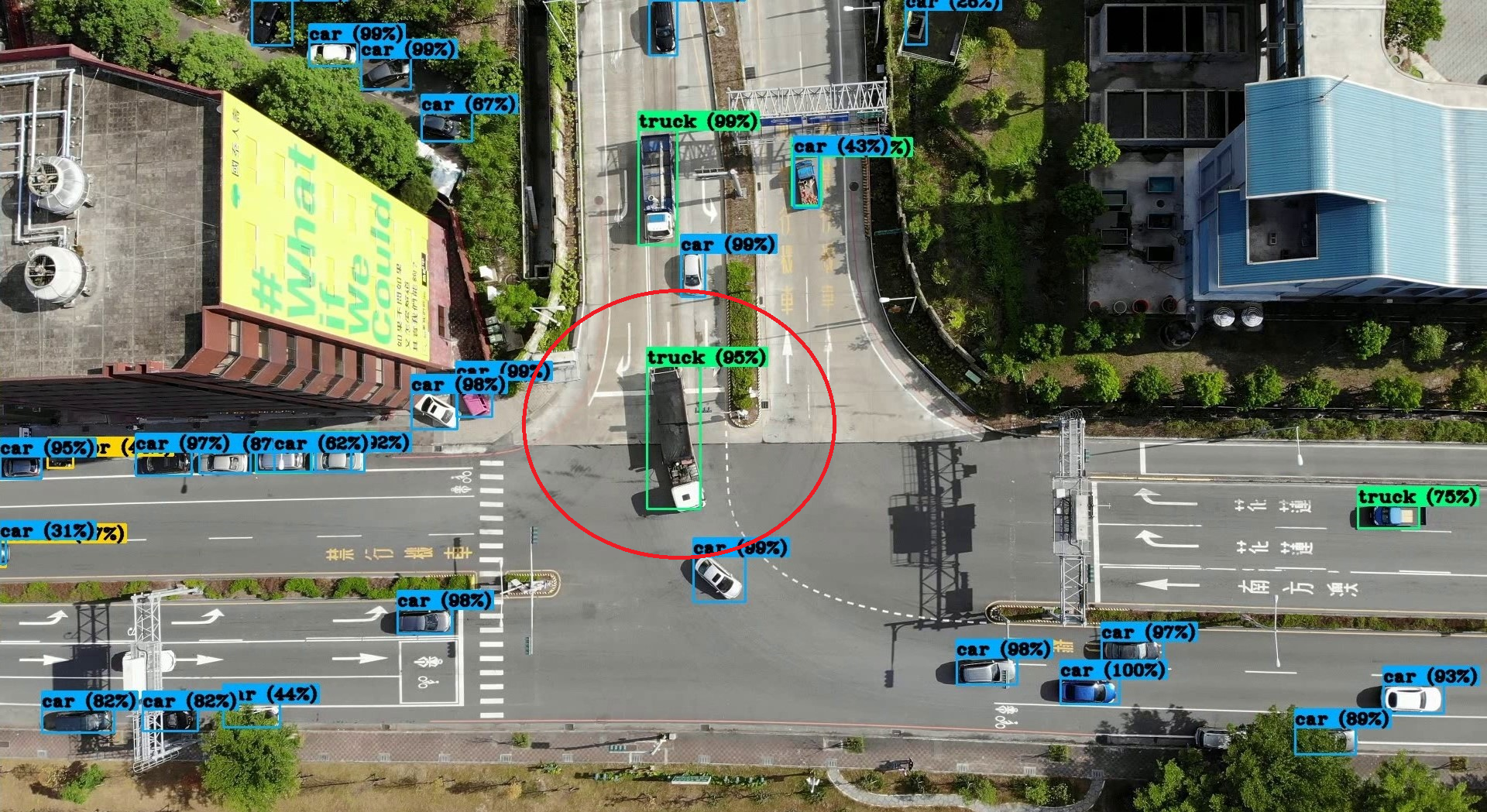}
}

\centerline{
  {\footnotesize (c)}
  \includegraphics[width=0.45\textwidth]{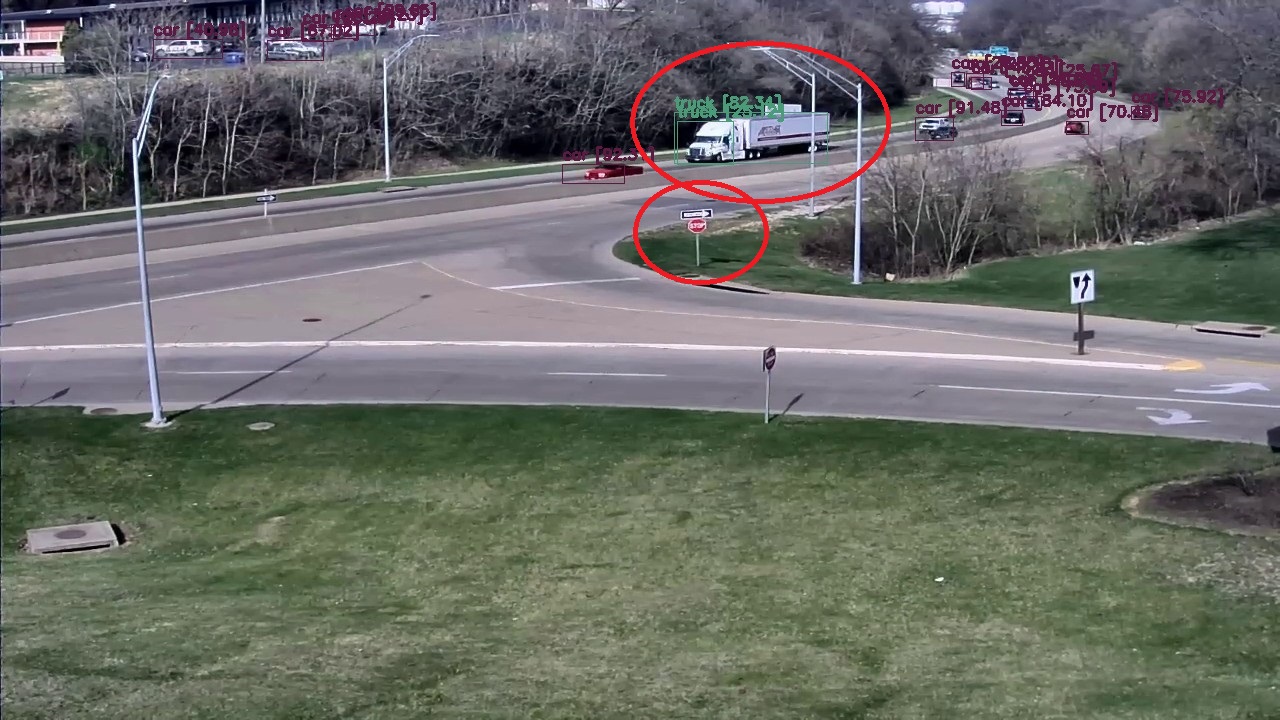}
   {\footnotesize (d)}
  \includegraphics[width=0.45\textwidth]{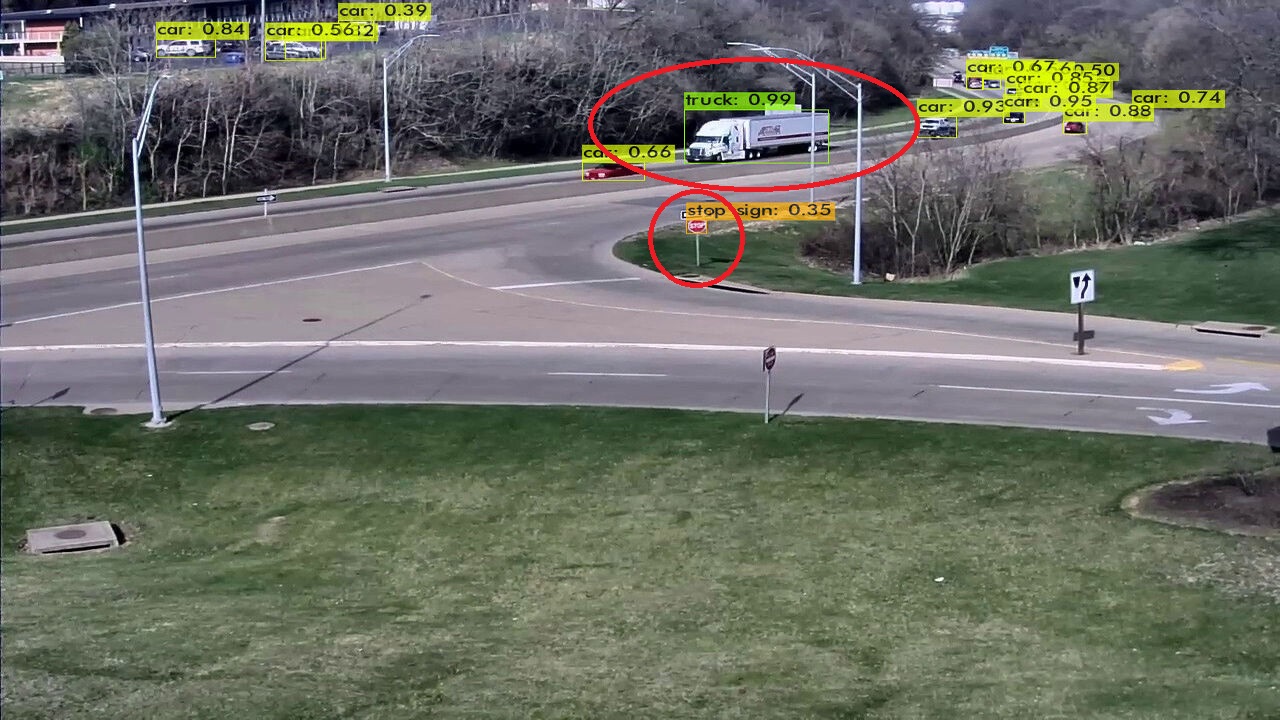}
  \vspace{-0.3cm}
} 
\caption{
Comparisons of object detection between YOLOv4 and our PRB-FPN.(a) and (b) are the results of our home pictures taken from aerial cameras in Suao, Taiwan; (c) and (d) are the results on the AI City Challenge~\cite{AICity20:CVPRW:2020}.
(a) : YOLOv4 $512\times512$,
(b) : PRB-FPN w/o $512\times512$.
(c) : YOLOv4 $512\times512$,
(d) : PRB-FPN w/o $512\times512$.}
\label{fig:YOLOv4Comparisons}
\end{figure*}

\begin{figure*}[t]
\centerline{
  {\footnotesize (a)}
   \includegraphics[width=0.45\textwidth]{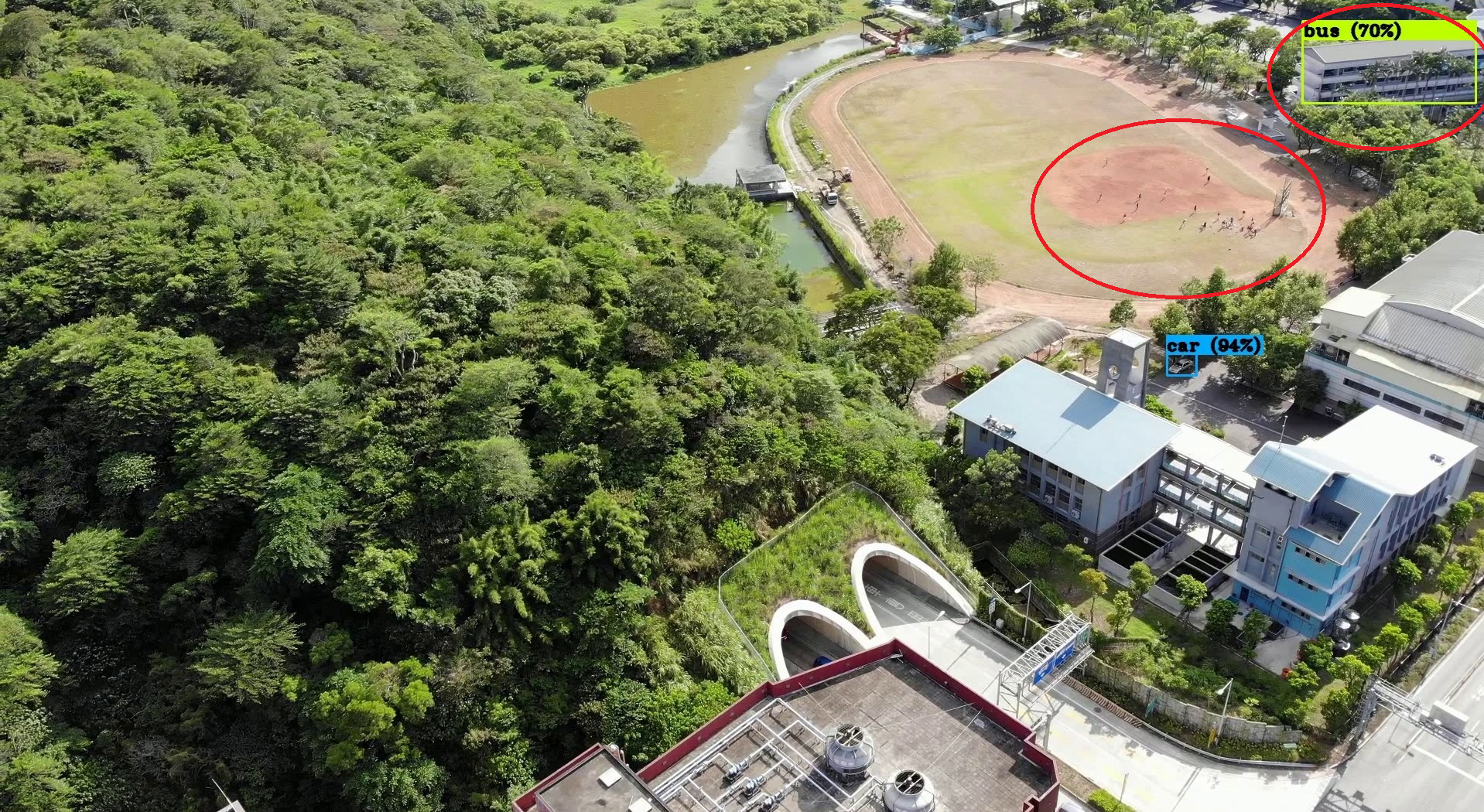}
   {\footnotesize (b)}
  \includegraphics[width=0.45\textwidth]{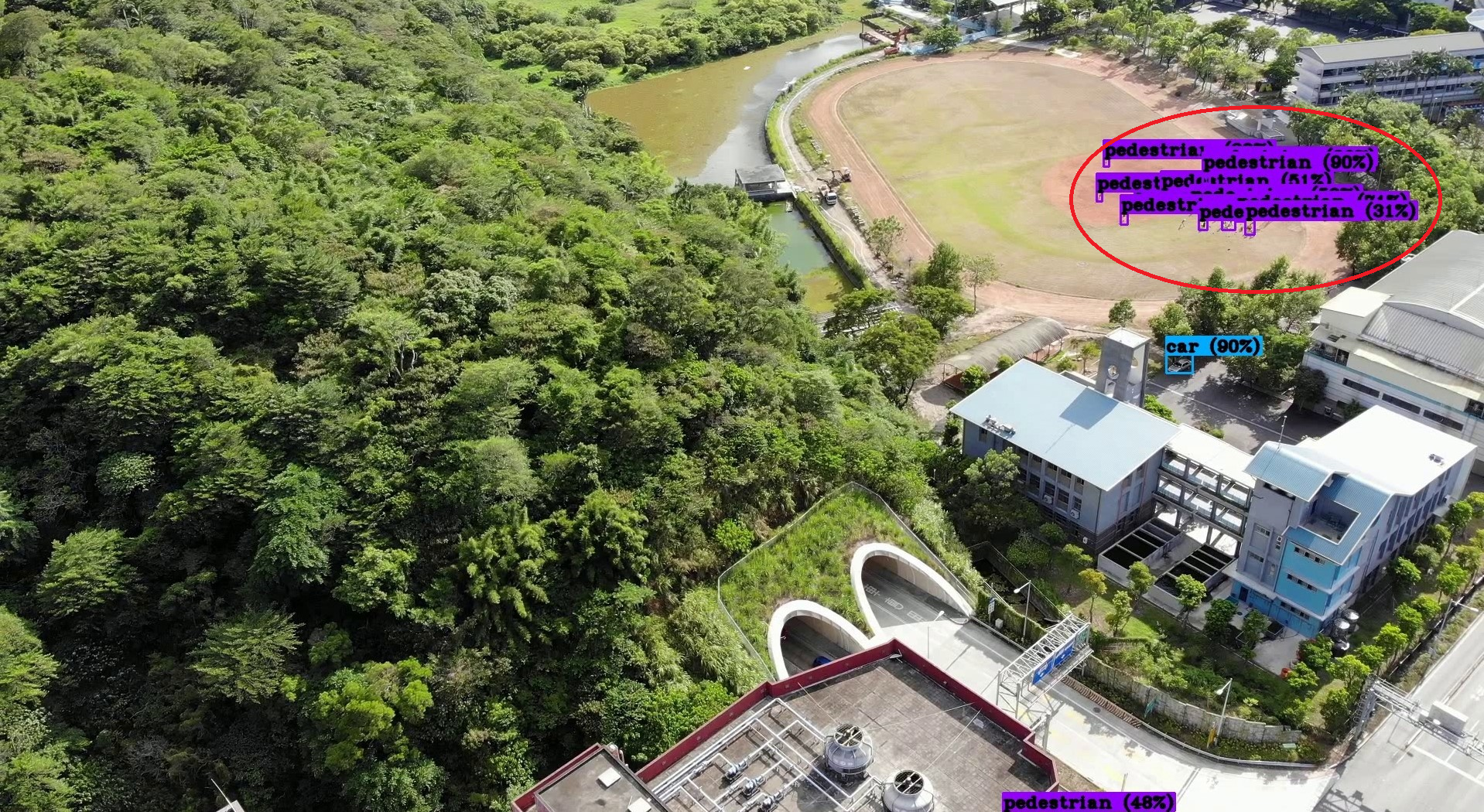}
\vspace{-0.3cm}
}  
\caption{
Comparisons of object detection between YOLOv4 and our PRB-FPN.(a) and (b) are the results of our home pictures taken from aerial cameras in Suao, Taiwan.
(a) : YOLOv4  $512\times512$,
(b) : PRB-FPN $512\times512$.}
\label{fig:YOLOv4ComparisonsAICity}
\end{figure*}

\begin{figure*}[t]
\centerline{
  {\footnotesize (a)}
   \includegraphics[width=0.45\textwidth]{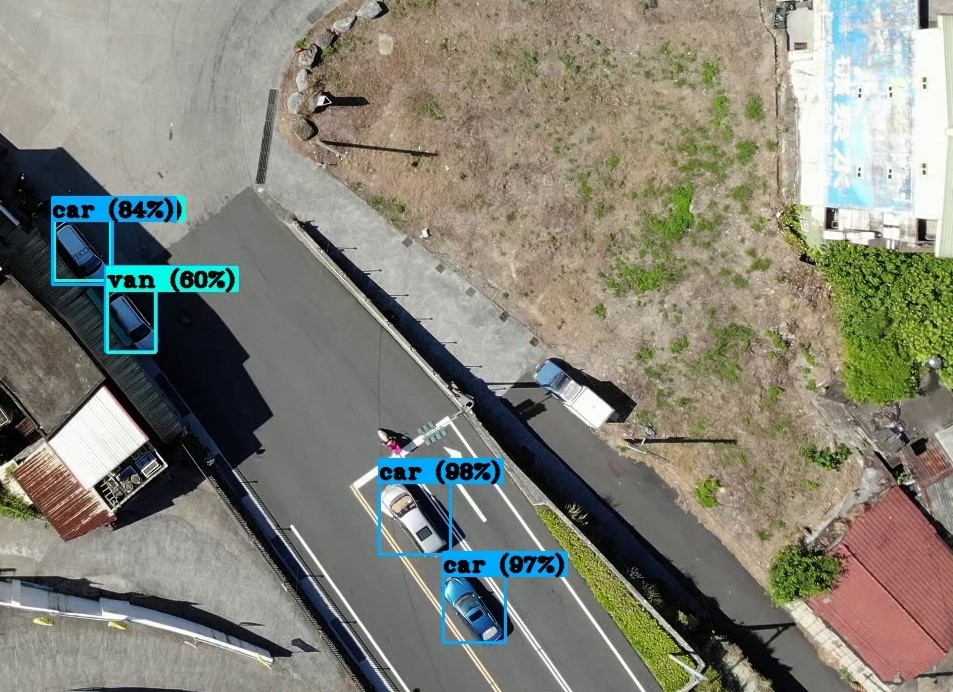}
   {\footnotesize (b)}
  \includegraphics[width=0.45\textwidth]{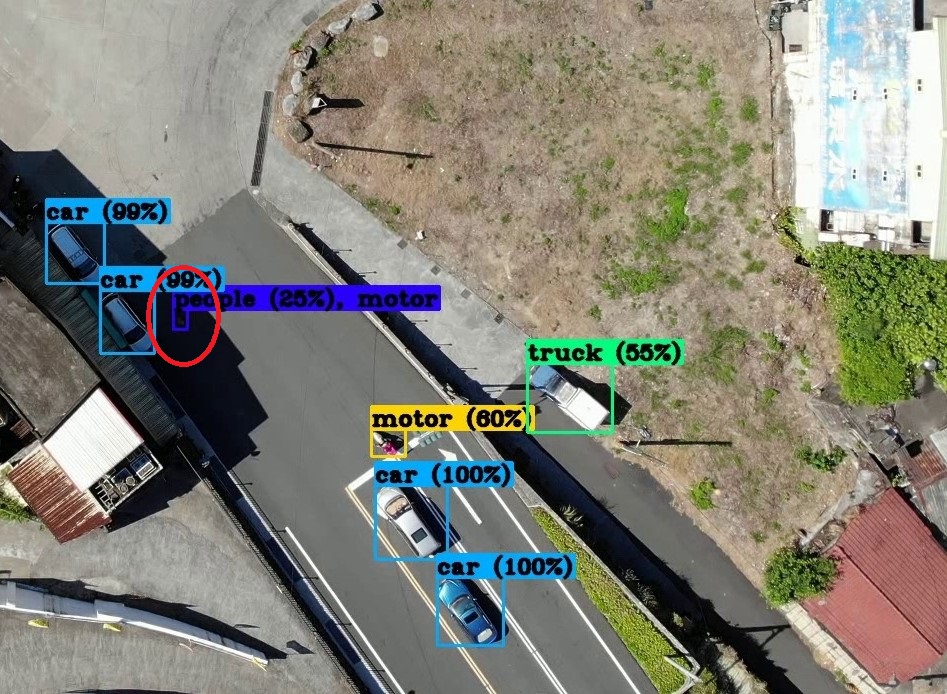}
\vspace{-0.3cm}
}  
\caption{
 Visualization of a failure case of PRB-FPN when compared with (a) YOLOv4 $512\times512$. (b) shows the result of PRB-FPN $512\times512$, where a false negative detection is shown in a red circle.
}
\label{fig:negativeResult}
\end{figure*}

\begin{figure}[t]
\centerline{
  \includegraphics[width=0.5\linewidth]{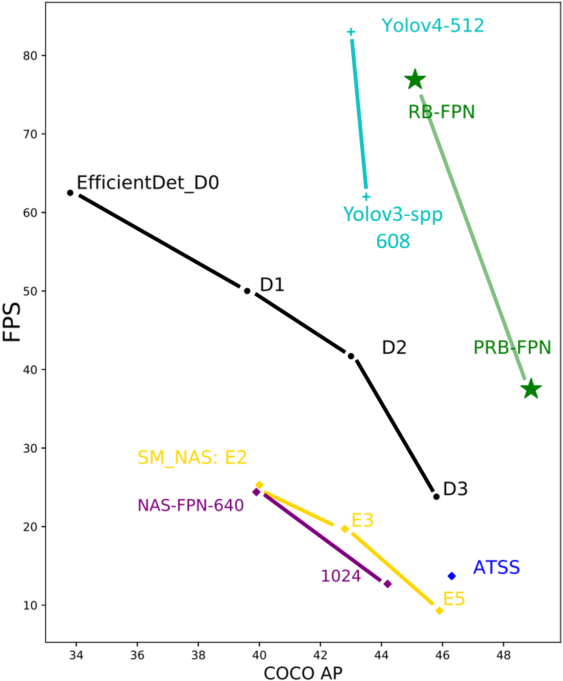}
  \vspace{-0.2cm}
}  
\caption{AP vs. inference time on MS COCO detection.
\vspace{-0.4cm}
}
\label{fig:APS:time}
\end{figure}

\begin{table}[t]
\scriptsize
\caption{Improvements by Parallel and Residual FPNs on UAVDT~\cite{UAVDT} Benchmark.
\vspace{-0.1cm}
}
\setlength\tabcolsep{2.5pt}
\begin{center}
\begin{tabular}{ccccc}
\toprule
\multicolumn{5}{c}{UAVDT-Benchmark-TestSet}                \\
\midrule
Methods                  & Backbone       & input size & AP    & FPS  \\
\midrule
Faster-RCNN      & VGG-16         & 1024x540   & 22.32 & 2.8  \\
R-FCN                    & ResNet-50      & 1024x540   & 34.35 & 4.7  \\
SSD              & VGG-16         & 512x512    & 33.62 & 41.56  \\
RON**                    & VGG-16         & 512x512    & 21.59 & 11.1 \\
RetinaNet       & ResNet-101-FPN & 512x512    & 33.95 & 25   \\
LRFNet          & VGG-16         & 512x512    & 37.81 & 91   \\
SpotNet            & Hourglass-104  & 512x512    & 52.8  & -    \\
CenterNet                & Hourglass-104  & 512x512    & 51.18 & -    \\
\midrule
BFM              & MobileNet-V2   & 512x512    & 29.7 & 113 \\
Re-CORE & MobileNet-V2   & 512x512    & 34.2 & 110 \\
PRB-FPN  & MobileNet-V2   & 512x512    & {\bf 65.47} & 75 \\
\midrule
Yolov4 with BFM     & CSPDarknet-53   & 512x512    & {\bf 64.52} & 30.5 \\
Yolov4 with Re-CORE & CSPDarknet-53   & 512x512    & {\bf 65.41} & 26.3 \\
Yolov4 with PRB-FPN  & CSPDarknet-53   & 512x512    & {\bf 76.55} & 19.2 \\
\bottomrule
\end{tabular}
\end{center}
\label{tbl:UAVDT}
\end{table}

\begin{table}[t]
\scriptsize
\caption{Ablation study of the Number of feature pyramidal layers for PRB with ResNet50 and RB with CSPDarknet53 on UAVDT~\cite{UAVDT} benchmark.
}
\centerline{
    \begin{tabular}{ccccccc}
    \toprule
    {} &{\scriptsize Number of FP layers} &{ } &Method &Backbone&FPS & mAP \\
    3&4&5&&&\\
    \midrule
    $\checkmark$ &&& PRB& ResNet-50& \textbf{31.26}& 70.71\\
    &$\checkmark$ && PRB& ResNet-50& 27.15& 72.32\\
    &&$\checkmark$ & PRB& ResNet-50& 22.30& 74.19\\
    \midrule
    $\checkmark$ &&& PRB& CSPDarknet-53& 19.2& 76.55\\
    &$\checkmark$&& PRB& CSPDarknet-53& 12.2& 77.82\\
    &&$\checkmark$& PRB& CSPDarknet-53& 4.7&\textbf{79.21}\\
    \bottomrule
    \end{tabular}
}
\label{tbl:PRB:UAVDT}
\end{table}

\begin{table*}[t]
\caption{Comparisons on the MS COCO test-dev set with SoTA models on nVidia Volta V100.
\vspace{-0.1cm}
}
\centering
\setlength\tabcolsep{2.5pt}
\centerline{
\begin{tabular}{llllllllll}
\toprule
Method                  & Backbone       & Input size     & FPS  & AP   & AP50 & AP75 & APS  & APM  & APL  \\
\toprule
YOLOv4~\cite{YOLOv4:arXiv:2020}                  & CSPDarknet-53~\cite{CSPNet}  & 512x512        & 83   & 43   & 64.9 & 46.5 & 24.3 & 46.1 & 55.2 \\
EfficientDet-D0~\cite{EfficientDet:CVPR:2020}         & Efficient-B0~\cite{EfficientNet:pmlr:2019}   & 512x512        & 97.0 & 33.8 & 52.2 & 35.8 & 12   & 38.3 & 51.2 \\
EfficientDet-D1~\cite{EfficientDet:CVPR:2020}         & Efficient-B1~\cite{EfficientNet:pmlr:2019}   & 640x640        & 74.0   & 39.6 & 58.6 & 42.3 & 17.9 & 44.3 & 56   \\
EfficientDet-D2~\cite{EfficientDet:CVPR:2020}         & Efficient-B2~\cite{EfficientNet:pmlr:2019}   & 768x768        & 57.0 & 43   & 62.3 & 46.2 & 22.5 & 47   & 58.4 \\
EfficientDet-D3~\cite{EfficientDet:CVPR:2020}         & Efficient-B3~\cite{EfficientNet:pmlr:2019}   & 896x896        & 36.0 & 47.5   & 66.2 & 51.5 & 27.9 & 51.4   & 62.0 \\
SM-NAS: E2~\cite{SMNAS:AAAI:2020}              &                & 800x600        & 25.3 & 40   & 58.2 & 43.4 & 21.1 & 42.4 & 51.7 \\
SM-NAS: E3~\cite{SMNAS:AAAI:2020}              &                & 800x600        & 19.7 & 42.8 & 61.2 & 46.5 & 23.5 & 45.5 & 55.6 \\
SM-NAS: E5~\cite{SMNAS:AAAI:2020}              &                & 1333x800        & 9.3 & 45.9 & 64.6 & 49.6 & 27.1 & 49.0 & 58.0 \\
NAS-FPN~\cite{NAS-FPN:CVPR:2019}                 & ResNet-50~\cite{ResNet}      & 640            & 24.4 & 39.9 &      &      &      &      &      \\
NAS-FPN~\cite{NAS-FPN:CVPR:2019}                 & ReNet-50~\cite{ResNet}       & 1024           & 12.7 & 44.2 &      &      &      &      &      \\
ATSS~\cite{ATSS:CVPR:2020}                    & ResNet-101~\cite{ResNet}     & 800x           & 17.5 & 43.6 & 62.1 & 47.4 & 26.1 & 47   & 53.6 \\                       
ATSS~\cite{ATSS:CVPR:2020}                    & ReNet-101~\cite{ResNet}      & 800x           & 13.7 & 46.3 & 64.7 & 50.4 & 27.7 & 49.8 & 58.4 \\
\midrule
RB-FPN [Ours]                & CSPDarknet-53~\cite{CSPNet}  & 512x512        &76.9 &45.1 &67.2 &48.2 &27.1 &48.5 &57\\
PRB-FPN [Ours]                & CSPDarknet-53~\cite{CSPNet}  & 800x800        &37.5 &48.9 &69.5 &55.9 &30.8 &55.9 &60.2\\
\bottomrule
\end{tabular}
}
\label{tbl:COCO:VoltaV}
\end{table*}

\begin{table*}[t]
\centerline{
\begin{tabular}{llllllllll}
\toprule
Method                  & Backbone       & Input size     & FPS  & AP   & AP50 & AP75 & APS  & APM  & APL  \\
\toprule
YOLOv7~\cite{wang_yolov7_2022}              & E-ELAN~\cite{wang_yolov7_2022}  & 640   &\textbf{161} &51.4 &69.7 &55.9 &31.8 &55.5 &65.0  \\
YOLOv7-X~\cite{wang_yolov7_2022}              & E-ELAN~\cite{wang_yolov7_2022}  & 640   &114 &53.1 &71.2 &57.8 &33.8 &57.1 &67.4  \\
YOLOR-CSP~\cite{yolor:2021:ArXiv}  &CSPNet~\cite{CSPNet} &640 &106 &51.1 &69.6 &55.7 &31.7 &55.3 &64.7 \\
YOLOR-CSP-X~\cite{yolor:2021:ArXiv}  &CSPNet~\cite{CSPNet} &640 &87 &53.0 &71.4 &57.9 &33.7 &57.1 &66.8 \\
EfficientDet-D4 ~\cite{EfficientDet:CVPR:2020} & Efficient-B4~\cite{EfficientNet:pmlr:2019} &1024 &24 &49.7 &68.4 &53.9 &- &- &- \\
YOLOv4-P6~\cite{scaledyolo:2021:CVPR}  &CSP-P6~\cite{scaledyolo:2021:CVPR} &1280 &32 &54.5 &72.6 &59.8 &36.8 &58.3 &65.9 \\
YOLOR-D6~\cite{yolor:2021:ArXiv} &CSPNet~\cite{CSPNet} &1280 &34 &56.5 &74.1 &61.9 &38.9 &60.4 &68.7 \\
YOLOv7-E6E~\cite{wang_yolov7_2022}  & E-ELAN~\cite{wang_yolov7_2022}             & 1280   &36 &56.8 &\textbf{74.4} &62.1 &\textbf{39.3} &\textbf{60.5} &69.0  \\
\midrule
PRB-FPN-CSP  [Ours]      & CSP-P5~\cite{scaledyolo:2021:CVPR}  & 640 &113 &51.8 &70.0 &56.7 &32.6 &55.5 &64.6  \\
PRB-FPN-MSP  [Ours]      & MSPNet~\cite{MSPNet:2022:ITS}  & 640 &94 &53.3 &71.1 &58.3 &34.1 &57.3 &66.2  \\
PRB-FPN-ELAN  [Ours]      & ELAN~\cite{wang_yolov7_2022}  & 640   &70 &52.5 &70.4 &57.2 &33.4 &56.2 &65.8  \\
PRB-FPN6 [Ours]   & E-ELAN~\cite{wang_yolov7_2022} & 1280   &31 &\textbf{56.9} &74.1 &\textbf{62.3} &39.0 &\textbf{60.5} &\textbf{70.0}  \\
\bottomrule
\end{tabular}
}
\caption{Comparison of state-of-the-art real-time object detectors.
}
\vspace{-2mm}
\label{tbl:COCO:VoltaV:realtime}
\end{table*}

\begin{table*}
\caption{Comparisons on the MS COCO test-dev set with SoTA models on nVidia Geforce Titan X. }
\vspace{-0.1cm}
\centerline{
\begin{tabular}{llllllllll}
\toprule
Method                  & Backbone       & Input size     & FPS  & AP   & AP50 & AP75 & APS  & APM  & APL  \\
\toprule
\midrule
\emph{two-stage:} & & & & & & & &\\
Faster R-CNN w/ FPN~\cite{FasterRCNN}          &VGGNet-16~\cite{VGG:ICLR:2015}      &$1000\times600$ &7.0 &21.9 &42.7 &- &- &- &-\\
Faster R-CNN w/ FPN~\cite{FasterRCNN}          &ResNet-101~\cite{ResNet}  &$1000\times600$ &6.0 &36.2 &59.1 &39.0 &18.2 &39.0 &48.2\\
OHEM++~\cite{OHEM:CVPR:2016}                   &VGGNet-16~\cite{VGG:ICLR:2015}      &$1000\times600$ &7.0 &25.5 &45.9 &26.1 &7.4 &27.7 &40.3\\
R-FCN~\cite{RFCN:NIPS:2016}                    &ResNet-101~\cite{ResNet}  &$1000\times600$ &9.0 &29.9 &51.9 &- &10.8 &32.8 &45.0\\
CoupleNet~\cite{CoupleNet:ICCV:2017}           &ResNet-101~\cite{ResNet}  &$1000\times600$ &8.2 &34.4 &54.8 &37.2 &13.4 &38.1 &50.8\\
Cascade R-CNN~\cite{CascadeRCNN:CVPR:2018}               &ResNet-101~\cite{ResNet}  &$1280\times800$ &7   &42.8 &62.1 &46.3 &23.7 &45.5 &55.2\\
Mask-RCNN~\cite{MaskRCNN}                      &ResNeXt-101~\cite{ResNet} &$\sim$1280x800  &3.3 &39.8 &62.3 &43.4 & 22.1 & 43.2 & 51.2 \\
Deformable R-FCN~\cite{DeformRFCN:ICCV:2017}   &ResNet-101~\cite{ResNet}  &$1000\times600$ &8 &34.5 &55 &- &14 &37.7 &50.3\\
Deformable R-FCN~\cite{DeformRFCN:ICCV:2017}   &Inc-Res-v2~\cite{DeformRFCN:ICCV:2017}  &$1000\times600$ &- &37.5 &58 &40.8 &19.4 &40.1 &52.5\\
Fitness-NMS~\cite{FitnessNMS:CVPR:2018}        &ResNet-101~\cite{ResNet}  &$1024\times1024$&5 &41.8 &60.9 &44.9 &21.5 &45.0 &57.5\\
SNIP~\cite{SNIP_Singh_2018_CVPR}               &DPN-98~\cite{SNIP_Singh_2018_CVPR}         &- &- &{\bf 45.7} &{\bf 67.3} &{\bf 51.1} &{\bf 29.3} &{\bf 48.8} &{\bf 57.1}\\
\midrule
\emph{one-stage low resolution:} & & & & & & & &\\
SSD~\cite{SSD}                        &VGGNet-16~\cite{VGG:ICLR:2015}      & 300x300        &43 &25.1 &43.1 &25.8 &6.6 &25.9 &41.4\\
RON~\cite{RON:CVPR:2017}                 & VGGNet-16~\cite{VGG:ICLR:2015}      & 384x384        & 15   & 27.4 & 49.5 & 27.1 & -   & - & - \\
DSSD~\cite{DSSD:Arxiv:2017}                  & ResNet-101~\cite{ResNet}      & 321x321        & 9.5   & 28.0 & 46.1 & 29.2 & 7.4   & 28.1 & 47.6 \\
RFBNet~\cite{RFBNet:ECCV:2018}               & VGGNet-16~\cite{VGG:ICLR:2015}      &300x300        &66.7   &30.3 &49.3 &31.8 &11.8 &31.9 &45.9 \\
YOLOv3~\cite{YOLOv3}                  &DarkNet-53~\cite{YOLOv3}      &416x416        &35   &31.0 &55.3 &32.3 &15.2 &33.2 &42.8 \\
PFPNet-R~\cite{PFPNet:ECCV:2018}      &VGG-16~\cite{VGG:ICLR:2015}      &320x320        &33   &31.8 &52.9 &33.6 &12 &35.3 &46.1 \\
RetinaNet~\cite{Retina}                & ResNet-101~\cite{ResNet}      &$\sim640\times400$        & 12.3   & 31.9 & 49.5 & 34.1 & 11.6   & 35.8 & 48.5 \\
RefineDet~\cite{RefineDet:CVPR:2018}                & VGGNet-16~\cite{VGG:ICLR:2015}      & 320x320        & 38.7   & 29.4 & 49.2 & 31.3 & 10.0   & 32.0 & 44.4 \\
RefineDet~\cite{RefineDet:CVPR:2018}                &ResNet-101~\cite{ResNet}      & 320x320        &-   &38.6 &59.9 &41.7 &21.1   &41.7 &52.3 \\
M2Det~\cite{M2Det:AAAI:2019}                &VGGNet-16~\cite{VGG:ICLR:2015}      & 320x320        &33.4 &33.5   &52.4 &35.6 &14.4 &37.6   &47.6\\
M2Det~\cite{M2Det:AAAI:2019}                 &ResNet-101~\cite{ResNet}      & 320x320        &21.7 &34.3   &53.5 &36.5 &14.8 &38.8   &47.9\\
LRFNet~\cite{LRF_Wang_2019_ICCV}                  & VGGNet-16~\cite{VGG:ICLR:2015}     & 300x300        &76.9  &32.0 &51.5 &33.8 &12.6  &34.9 &47.0 \\
LRFNet~\cite{LRF_Wang_2019_ICCV}                  & ResNet-101~\cite{ResNet}     &300x300        &52.6   &34.3 &54.1 &36.6 &13.2 &38.2 &50.7 \\
\midrule
\emph{one-stage high resolution:} & & & & & & & &\\
LRFNet~\cite{LRF_Wang_2019_ICCV}                  & VGGNet-16~\cite{VGG:ICLR:2015}     & 512x512        & 38   & 36.2 & 56.6 & 38.7 & 19   & 39.9 & 48.8 \\
LRFNet~\cite{LRF_Wang_2019_ICCV}                  & ResNet-101~\cite{ResNet}     & 512x512        & 31   & 37.3 & 58.5 & 39.7 & 19.7 & 42.8 & 50.1 \\
EFIP~\cite{EFIP:CVPR:2019}                    & VGGNet-16~\cite{VGG:ICLR:2015}      & 512x512        & 34   & 34.6 & 55.8 & 36.8 & 18.3 & 38.2 & 47.1 \\
RFBNet~\cite{RFBNet:ECCV:2018}                  & VGGNet-16~\cite{VGG:ICLR:2015}      & 512x512        & 33   & 33.8 & 54.2 & 35.9 & 16.2 & 37.1 & 47.4 \\
RFBNet-E~\cite{RFBNet:ECCV:2018}                 & VGGNet-16~\cite{VGG:ICLR:2015}      & 512x512        & 30   & 34.4 & 55.7 & 36.4 & 17.6 & 37   & 47.6 \\
SSD~\cite{SSD}        & ResNet101~\cite{ResNet}      & 513x513        & 31.3 & 31.2 & 50.4 & 33.3 & 10.2 & 34.5 & 49.8 \\
SSD~\cite{SSD}        & VGGNet-16~\cite{VGG:ICLR:2015}      & 512x512        & 22   & 28.8 & 48.5 & 30.3 & 10.9 & 31.8 & 43.5 \\
DSSD~\cite{DSSD:Arxiv:2017}        &ResNet101~\cite{ResNet}      & 513x513        & 5.5 & 33.2 & 53.3 & 35.2 & 13.0 & 35.4 & 51.1 \\
YOLOv2~\cite{YOLOv2}   & DarkNet-19~\cite{YOLOv2}     & 544x544        & 40   & 21.6 & 44   & 19.2 & 5    & 22.4 & 35.5 \\
YOLOv4~\cite{YOLOv4:arXiv:2020}  & CSPDarknet-53~\cite{CSPNet}  & 512x512        & 31   & 43   & 64.9 & 46.5 & 24.3 & 46.1 & 55.2 \\
YOLOv3-SPP~\cite{YOLOv3} & DarkNet-53~\cite{YOLOv3}     & 608x608        & 19.8 & 36.2 & 60.6 & 38.2 & 20.6 & 37.4 & 46.1 \\
YOLOv3-SPP~\cite{YOLOv3} & DarkNet-53~\cite{YOLOv3}     & 608x608        & 19.8 & 36.2 & 60.6 & 38.2 & 20.6 & 37.4 & 46.1 \\
RefineDet\cite{RefineDet:CVPR:2018} & VGGNet-16~\cite{VGG:ICLR:2015}      & 512x512        & 22.3 & 33   & 54.5 & 35.5 & 16.3 & 36.3 & 44.3 \\
CornerNet~\cite{coreNet:ECCV:2018} &Hourglass~\cite{coreNet:ECCV:2018}      & 512x512        &4.4 &40.5 &57.8 &45.3 &20.8 &44.8 &56.7 \\
M2Det~\cite{M2Det:AAAI:2019}     & VGGNet-16~\cite{VGG:ICLR:2015}      & 512x512        & 18   & 37.6 & 56.6 & 40.5 & 18.4 & 43.4 & 51.2 \\
M2Det~\cite{M2Det:AAAI:2019}   & ResNet-101~\cite{ResNet}     & 512x512        & 15.8 & 38.8 & 59.4 & 41.7 & 20.5 & 43.9 & 53.4 \\
RetinaNet~\cite{Retina}               & ResNet-50~\cite{ResNet}  & $\sim$832x500  & 13.9 & 32.5 & 50.9 & 34.8 & 13.9 & 35.8 & 46.7 \\
RetinaNet~\cite{Retina}               & ResNet-101~\cite{ResNet} & $\sim$832x500  & 11   & 34.4 & 55.7 & 36.8 & 14.7 & 37.1 & 47.4 \\
RetinaNet+AP-Loss~\cite{Retina}       & ResNet-101~\cite{ResNet} & 512x512        & 11   & 37.4 & 58.6 & 40.5 & 17.3 & 40.8 & 51.9 \\
ACoupleNet~\cite{AttentionCoupleNet:TIP:2019}           & ResNet-101~\cite{ResNet} & 600x1000 & -    & 35.4 & 55.7 & 37.6 & 13.2 & 38.6 & 52.5 \\
SAFNet~\cite{SAFNet:TIP:2020}           & ResNet-101~\cite{ResNet} & 768x768 & -    & 39.2 & 60.6 & 42.3 & 20.2 & 44.2 & 52.6 \\
Cascade R-CNN~\cite{CascadeRCNN:CVPR:2018}           & ResNet-101~\cite{ResNet} & $\sim$1280x800 & 7    & 42.8 & 62.1 & 46.3 & 23.7 & 45.5 & 55.2 \\
FoveaBox~\cite{FoveaBox:TIP:2020}           & ResNeXt-101~\cite{ResNet} & 800x800 & -    & 42.3 & 62.9 & 45.4 & 25.3 & 46.8 & 55.0 \\
AB+FSAF~\cite{AB:FSAF:CVPR:2019}                 & ResNet-101~\cite{ResNet}     & 800            & 5.6  & 40.9 & 61.5 & 44   & 24   & 44.2 & 51.3 \\
AB+FSAF~\cite{AB:FSAF:CVPR:2019}                 & ResNeXt-101~\cite{ResNet}    & 800            & 2.8  & 42.9 & 63.8 & 46.3 & 26.6 & 46.2 & 52.7 \\
\midrule
RB-FPN [Ours] & ResNet-50~\cite{ResNet}  & 512x512  & 32.1 & 44.3 & 65.1 & 48.2 & 25.1 & 47.3 & 56.8 \\
PRB-FPN [Ours] & ResNet-50~\cite{ResNet} & 800x800 & 15.9 & 46.1 & 67.3 & 49.9 & 28.5 & 49.3 & 59.4 \\
RB-FPN [Ours] & CSPDarknet-53~\cite{CSPNet}  & 512x512        &27.3 &\textbf{45.1} &\textbf{67.2} &\textbf{48.2} &\textbf{27.1} &\textbf{48.5} &\textbf{57}\\
PRB-FPN [Ours]                & CSPDarknet-53~\cite{CSPNet}  & 800x800        &11.6 &\textbf{48.9} &\textbf{69.5} &\textbf{55.9} &\textbf{30.8} &\textbf{55.9} &\textbf{60.2}\\
\bottomrule
\end{tabular}
}
\label{tbl:COCO:TitanX}
\end{table*}

\subsection{BFM Accuracy Improvements}
\label{sec:eval:BFM}

Since the BFM in PRB-FPN is designed to detect both small and large objects, we evaluate the effectiveness of BFM on object detection based on the MS COCO dataset across four backbones, namely PeLee \cite{wang2018pelee}, DarkNet-53 \cite{YOLOv3}, VGG16 \cite{VGG:ICLR:2015}, and DenseNet \cite{huang2017densely}), to evaluate the generalization capability of BFM. Table~\ref{tbl:backbones} shows this BFM ablation study results. Observe that the BFM computational load is very light and can be ignored for all backbones. Also observe the generalizability of BFM in maintaining high AP across these backbones for detecting different object sizes. Table~\ref{tbl:backbones} also shows another important observation that BFM can improve the detection accuracy of a shallower backbone more than deeper backbone. Specifically, the improvements on AP50 with BFM for DarkNet \cite{YOLOv3}, Pelee \cite{wang2018pelee}, and DenseNet \cite{huang2017densely} are 6.5\%, 3.5\%, and 0.2\%, respectively. This indicates that BFM improves the detection of large objects better than smaller objects. Thus, BFM can provide a good solution for applications that demands the detection of arbitrary-sized objects.

In addition to the ablation study of our BFM method among different backbones, Table~\ref{tbl:biFPNCompare} shows the comparisons against other SoTA bifusion methods in terms of accuracy and efficiency.  When the backbones \texttt{ResNet-50} and EfficientNet are adopted, our BFM method outperforms EfficientDet-D0~\cite{EfficientDet:CVPR:2020} and NAS-FPN~\cite{NAS-FPN:CVPR:2019}. As for the bidirectional FPN-BPN~\cite{Wu2018SingleShotBP}, their convolutions with stride 2 for down-sampling result in lower accuracy in small object detection. In addition, their de-convolution for upsampling results in lower efficiency for object detection. 

\subsection{BFM with Re-CORE Accuracy Improvements}
\label{sec:eval:ReCORE}

We evaluate the effectiveness of BFM with Re-CORE for Residual Bi-Fusion object detection. Table \ref{tbl:ReCORE:BFM} shows the ablation study of the PRB-FPN {\em vs.} YOLOv3 and YOLOv4 with or without BFM Re-CORE. As a result, PRB-FPN outperforms YOLOv3 in all categories. Note that the frame rates with or without BFM are very similar. For input size $512\times512$, YOLOv3 with BFM also outperforms YOLOv3 alone on all categories. BFM improves the detection of small objects significantly, with an increasing trend as the input image size increases. On the contrary, improvements on the large objects have a decreasing trend as the input size increases.

Fig.~\ref{fig:ablationstudyvisulize} shows the ablation study comparisons of object detectors regarding the effects of BFM and Re-CORE modules on a selected image from \texttt{COCO-test-dev}. Fig.~\ref{fig:ablationstudyvisulize}(b) shows detections obtained by YOLOv3.  Fig.~\ref{fig:ablationstudyvisulize}(c) and (d) show detections of YOLOv3 with BF and Re-CORE modules, respectively.  
Fig.~\ref{fig:ablationstudyvisulize}(e) shows detections of the proposed PRB-FPN.  In comparison, Fig.~\ref{fig:ablationstudyvisulize}(f) shows detections obtained by M2Det~\cite{M2Det:AAAI:2019}. Observe clearly that the proposed PRB-FRN outperforms YOLOv3, YOLOv4 and M2Det.  

Fig.~\ref{fig:SOD} shows the comparisons of object detection against LRFNet~\cite{LRF_Wang_2019_ICCV} on a $1024\times540$ image from the UAVDT17 benchmark~\cite{UAVDT}. Note that the black masks in Fig.~\ref{fig:SOD} come with the original images in UAVDT for privacy protection.  LRFNet fails to detect the tiny far-away vehicles from the camera view, while PRB-FPN can successfully detect most of them. 

\subsection{PRB-FPN vs the Original FPN}
\label{sec:eval:Orig:FPN}

We compare the performance the proposed PRB-FPN {\em vs.} the original FPN on the UAVDT~\cite{UAVDT} benchmark. Performance evaluation on the MS COCO dataset is omitted, as it contains very few samples of small objects. Table~\ref{tbl:UAVDT} shows the performance comparisons with and without the proposal {\em parallel} or the {\em residual bi-fusion} modules. We adopted two backbones, namely MobileNet-V2~\cite{MobileNet-V2} and \texttt{CSPDarknet-53}~\cite{CSPNet} in this evaluation. \texttt{CSPDarknet-53} was created in our previous framework and is now adopted in YOLOv4. The baseline of FPN is the single bi-fusion module adopted in SoTA bi-fusion methods \cite{conf/icann/SchererMB10,PFPNet:ECCV:2018,Woo2019GatedBF}.

When the MobileNet-V2 backbone is used, accuracy of the baseline method ({\em i.e.} single bi-fusion module) is 29.7\%. In comparison, as the Re-CORE module is added, the accuracy improves from 29.7\% to 34.2\%. 
However, the score is still lower than LRFNet~\cite{LRF_Wang_2019_ICCV}, SpotNet~\cite{SpotNet}, and CenterNet~\cite{CenterNet}, since MobileNet-V2 is a very lightweight network.    
Finally, after the {\em parallel} FP bi-fusion design is included, the accuracy improves significantly from 34.3\% to 65.47\%, which outperform all comparison methods.  
Note that our PRB-FPN achieves double the amount of accuracy and triple amount of efficiency over RetinaNet~\cite{Retina}.  

Evaluation with the \texttt{CSPDarknet-53} backbone is shown in the last three rows of Table~\ref{tbl:UAVDT}.  Accuracy improvement is significant from 64.52 \% to 76.55 \%.  These evaluations suggest that our {\em parallel} FP bi-fusion design is general for accuracy improvement.  Table~\ref{tbl:PRB:UAVDT} shows the effects on accuracy and FPS of the number of feature pyramid layers. Two backbones, $i.e.$, \texttt{ResNet-50} and \texttt{CSPDarknet-53} were adopted in this evaluation. Clearly, with more FP layers, higher accuracy can be obtained for both backbones. Also the use of more FP layers results in lower FPS.  

Fig.~\ref{fig:YOLOv4Comparisons} shows visual comparisons of object detection between YOLOv4 and PRB-FPN. YOLOv4 cannot detect both large and small objects well enough. The enlargement of input image to detect small objects in YOLOv4 often fails in detecting large objects and undesired inefficiency. In Fig.~\ref{fig:YOLOv4Comparisons}(a), the cargo truck was missed by YOLOv4 but successfully detected by our PRB-FPN.  In addition, YOLOv4 often detects a large object as several small ones.  As shown  Fig.~\ref{fig:YOLOv4Comparisons}(c), a truck was detected as two cars, the PRB-FPN can detect it successfully in Fig.~\ref{fig:YOLOv4Comparisons}(d). Note that the stop sign in Fig.~\ref{fig:YOLOv4Comparisons}(c) was missed by YOLOv4. Without the enlargement, YOLOv4 will further miss-detect or incorrectly classify small objects. For example, the small persons on the playground (highlighted in a red circle) in Fig.~\ref{fig:YOLOv4ComparisonsAICity}(a) were missed by YOLO 4, while PRB-FPN can successfully detect them in Fig.~\ref{fig:YOLOv4ComparisonsAICity}(b). Also, in Fig.~\ref{fig:YOLOv4ComparisonsAICity}(a), a small building was wrongly detected as a bus by YOLOv4. High detection rate and high recall rate for small objects are the major characteristics of our PRB model.  Fig. ~\ref{fig:negativeResult}(b) shows a failure case of our method. A false detection of a pedestrian (shown with a red circle) occurred due to the dark background. In comparison, YOLOv4 detection results in Fig.~\ref{fig:negativeResult}(a) is visually better without a false detection, however this is due to the weakness of YOLOv4 in identifying small objects.  Such weakness of YOLOv4 can explain the miss detection of small vehicles and trucks in Fig. ~\ref{fig:negativeResult}(a). 


\subsection{Comparisons with State-of-The-Art Models}
\label{sec:eval:SOTA}

Tables \ref{tbl:COCO:VoltaV} and \ref{tbl:COCO:TitanX}  compare the PRB-FPN with and without parallelization design against other SoTA object detectors in terms of accuracy and efficiency.  Here experiments are conducted on two backbones of \texttt{CSPDarknet-53}~\cite{CSPNet} and \texttt{ResNet-50}~\cite{ResNet} for the performance comparison of PRB-FPN.  
To make fair comparisons, we did not evaluate the performance of anchor-free methods as their efficiency scores were not reported. Inference time is calculated as the average of execution time of the network with Non-Maximum Suppression (NMS) from 999 random images.

Fig.~\ref{fig:APS:time} plots the inference time vs. {\em mean Average Precision small} (APs) \cite{COCO:ECCV:2014} for many evaluated models, where the PRB-FPN was tested on nVidia Titan X. Observe that PRB-FPN (green curve in Fig.~\ref{fig:APS:time}) achieves outstanding speed-accuracy performance compared to other SoTA models. We highlight two advantages of the PRB-FPN: (1) the parallel bi-fusion design for multi-scale feature extraction can detect both small and large objects at the same time with higher accuracy , and (2) the fusion module can  effectively  fuse both  deep  and  shallow  feature  layers  in  parallel  for  fast and accurate one-shot object detection, specially for small objects. Observe that PRB-FPN outperforms the other SoTA one-stage object detectors (YOLOv4, YOLOv3, EfficientDet, ATSS, SM\_NAS, and NAS-FPN), when taking both the accuracy and speed into account. 
\subsection{Comparisons with State-of-The-Art real-time Models}
\label{exp:realtime:SOTA}

As illustrated in Table~\ref{tbl:COCO:VoltaV:realtime}, a comparative analysis was conducted between the PRB-FPN equipped with various backbones~\cite{CSPNet,MSPNet:2022:ITS,wang_yolov7_2022,MSPNet:2021:ICIP} and the existing SoTA methods~\cite{yolor:2021:ArXiv,scaledyolo:2021:CVPR,wang_yolov7_2022,EfficientDet:CVPR:2020}. Noticeably, PRB-FPN with a scaled P6 backbone surpasses all contemporary methods, thereby establishing a new SoTA on the real-time object detection COCO benchmark~\cite{COCO:ECCV:2014}. Furthermore, the P5 variant of PRB-FPN, incorporating MSPNet~\cite{MSPNet:2021:ICIP,MSPNet:2022:ITS}, also exhibits superior performance over the YOLOv7-X~\cite{wang_yolov7_2022} equipped with an ELAN~\cite{wang_yolov7_2022} backbone, while maintaining a comparable computational cost.

\begin{figure*}[t]
\centerline{
    {\footnotesize (a)}
    \includegraphics[width=.47\textwidth]{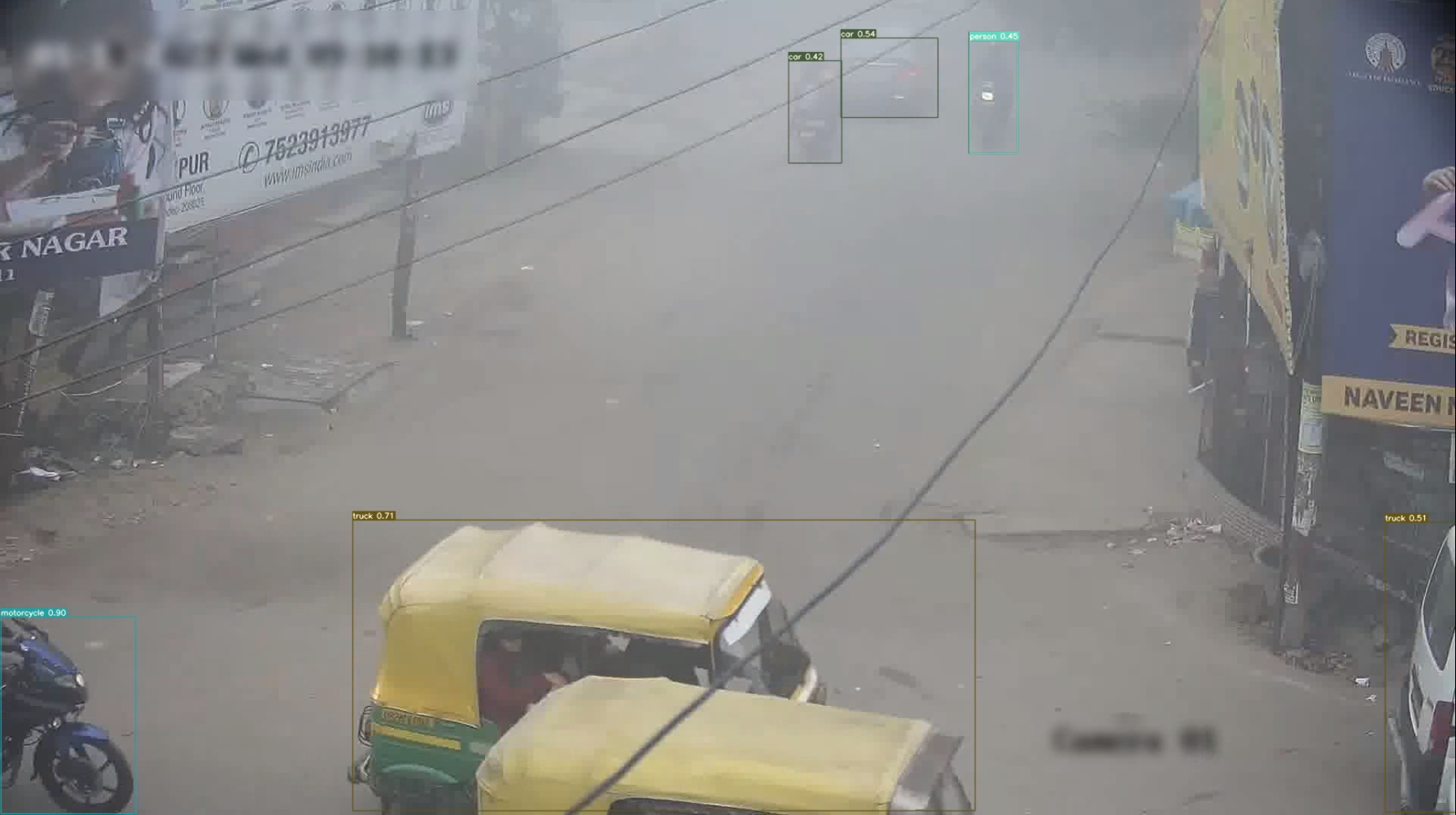}
    {\footnotesize(b)}
    \includegraphics[width=.47\textwidth]{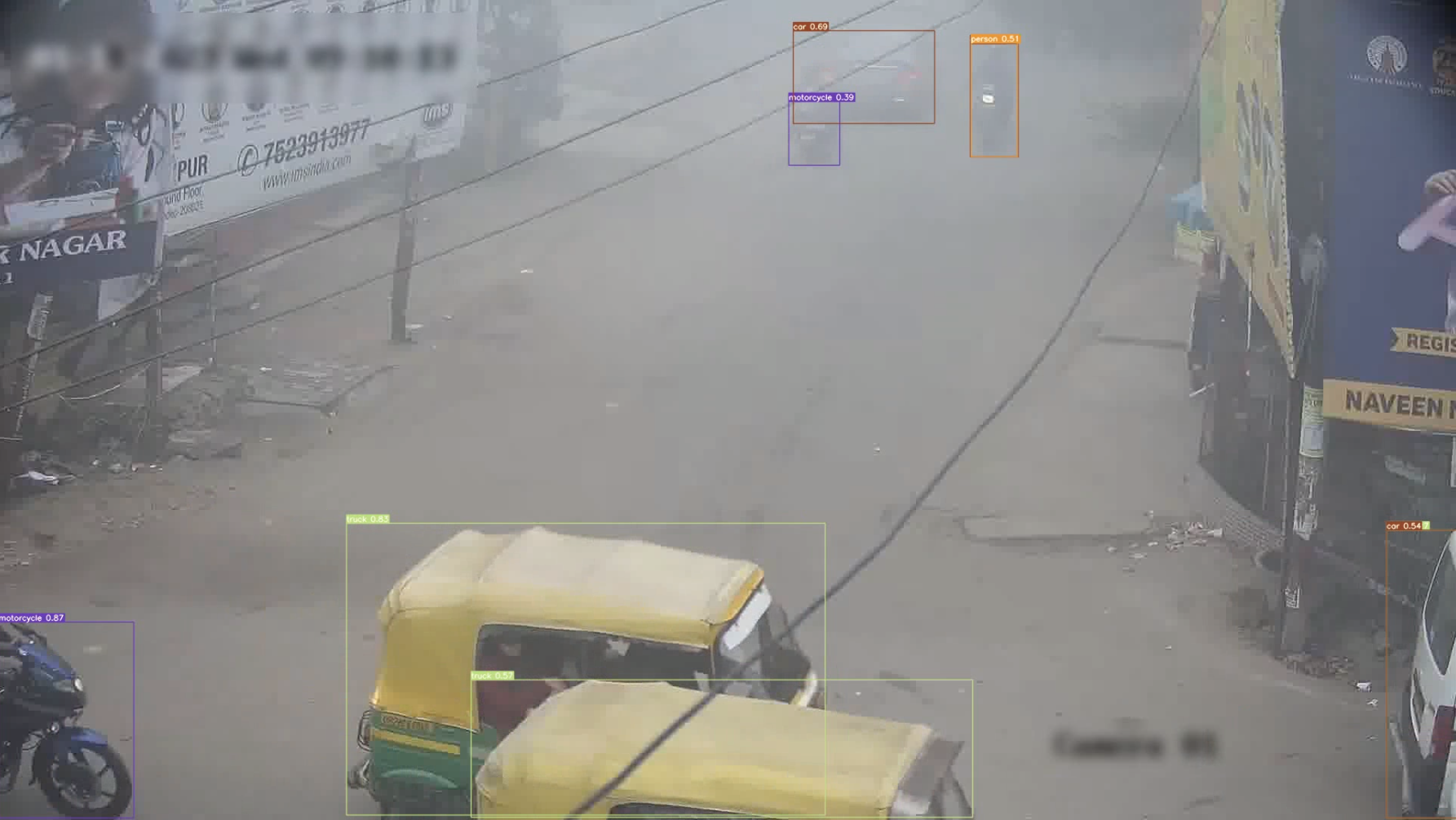}
}
\caption{Comparisons of object detection between YOLOv7-E6E~\cite{wang_yolov7_2022} and our PRB-FPN6. (a) shows the detection results obtained using YOLOv7-E6E~\cite{wang_yolov7_2022}, which were unable to accurately detect a motor in a foggy scene. (b) shows the detection results obtained using our proposed PRB-FPN6, which successfully recognized the objects in the foggy scene. These results demonstrate the superior performance and robustness of our approach compared to YOLOv7-E6E~\cite{wang_yolov7_2022} in challenging environments such as fog.
}
\label{fig:teaser}
\end{figure*}

A visual comparison between the YOLOv7-E6E~\cite{wang_yolov7_2022} detector and the PRN-FPN6 is presented in Figure~\ref{fig:teaser}. It demonstrates the enhanced capability of our proposed PRN-FPN6 in localizing objects, particularly those that are small or heavily occluded.

\section{Conclusions}

We present a new PRB-FPN model that can effectively fuse deep and shallow pyramid features for fast and accurate object detection. 
Our novel bi-directional residual FP design enables easy training and integration with popular backbones. The proposed bottom-up fusion improves the detection of both small and large objects.  
Extensive evaluations  show that PRB-FPN outperforms other bi-directional methods and SoTA one-stage methods, in terms of accuracy and speed.

{\bf Future work} includes the development of anchor-free methods that can avoid handcrafted anchors, which might further improves detection accuracy. Finally, Network Architecture Search (NAS) can potentially be adopted to find the better architecture, considering both the backbone and FP structures.


\section{Acknowledgement}

The authors sincerely appreciate Mr. Yuwei Chen for proofreading and improving the English writing of this paper. We thank to National Center for High-performance Computing (NCHC) for providing computational and storage resources. We sincerely appreciate \href{mailto:1155142198@link.cuhk.edu.hk}{Wing-Kit, Chan} (\emph{CUHK AIST major, year 4 in 2022-2023.}) and \href{mailto:1155157271@link.cuhk.edu.hk}{Hao-Yuan, Yue} (\emph{CUHK CS major, year 3 in 2022-23.}) for expanding our PRBNet in the YOLO PyTorch family~\cite{YOLOv4:arXiv:2020,yolov5:github:2021,wang_yolov7_2022} and oriented object detection~\cite{hukaixuan_yolov5obb_2022}.

\bibliographystyle{unsrt}  
\bibliography{ref}  

\end{document}